\documentclass[journal]{IEEEtran}
\usepackage{url}
\usepackage{times}
\usepackage{epsfig}
\usepackage{colortbl}
\usepackage{graphicx}
\usepackage{array}
\usepackage{amsmath}
\usepackage{amssymb}
\usepackage{multirow}
\usepackage{multicol}
\usepackage{dsfont}
\usepackage{caption}
\usepackage{color}
\usepackage{booktabs}
\usepackage{nth}
\usepackage{float}
\usepackage{bbding}
\usepackage{cite}
\usepackage{xspace}
\usepackage{tabularx}
\usepackage{setspace}
\usepackage{bbm}

\usepackage{placeins}
\restylefloat{table}
\usepackage{bm}
\newcommand{\PreserveBackslash}[1]{\let\temp=\\#1\let\\=\temp}
\newcolumntype{C}[1]{>{\PreserveBackslash\centering}p{#1}}
\newcolumntype{R}[1]{>{\PreserveBackslash\raggedleft}p{#1}}
\newcolumntype{L}[1]{>{\PreserveBackslash\raggedright}p{#1}}
\newcolumntype{?}{!{\vrule width 0.6pt}}

\usepackage{xspace}
\newcommand{\ie}{{\emph{i.e.}}\xspace}
\newcommand{\eg}{{\emph{e.g.}}\xspace}

\newcommand{\etal}{{\emph{et al.}}\xspace}

\usepackage{bbding}
\usepackage{pifont}
\usepackage{color,xcolor}
\usepackage{stfloats}
\usepackage{wasysym}
\usepackage{tabu}
\newcommand{\cmark}{\ding{51}\xspace}%
\newcommand{\xmark}{\ding{55}\xspace}%
\newcommand{\xmarkg}{\textcolor{lightgray}{\ding{55}}\xspace}%
\usepackage[pagebackref=false,breaklinks=true,letterpaper=true,colorlinks,linkcolor=red,bookmarks=false]{hyperref}

\ifCLASSINFOpdf
\else
\fi
\begin{document}
%
\title{Prototype Adaption and Projection for Few- and Zero-shot 3D Point Cloud Semantic Segmentation}
%
%
%

\author{Shuting He,
        Xudong Jiang~\IEEEmembership{IEEE Fellow},
        Wei Jiang,
        Henghui Ding
\thanks{Shuting He and Wei Jiang are with the State Key Laboratory of
Industrial Control Technology, College of Control Science and Engineering,
Zhejiang University, Hangzhou 310027, China.}

\thanks{Xudong Jiang and Henghui Ding are with Nanyang Technological University (NTU), Singapore 639798
(e-mail:exdjiang@ntu.edu.sg, henghui.ding@gmail.com).}
\thanks{(Corresponding author: Henghui Ding.)}
}

%
%

\markboth{IEEE Transactions on Image Processing}%
{Shell \MakeLowercase{\textit{et al.}}: Bare Demo of IEEEtran.cls for IEEE Journals}
%



\maketitle

\begin{abstract}
In this work, we address the challenging task of few-shot and zero-shot 3D point cloud semantic segmentation. The success of few-shot semantic segmentation in 2D computer vision is mainly driven by the pre-training on large-scale datasets like imagenet. The feature extractor pre-trained on large-scale 2D datasets greatly helps the 2D few-shot learning. However, the development of 3D deep learning is hindered by the limited volume and instance modality of datasets due to the significant cost of 3D data collection and annotation. This results in less representative features and large intra-class feature variation for few-shot 3D point cloud segmentation. As a consequence, directly extending existing popular prototypical methods of 2D few-shot classification/segmentation into 3D point cloud segmentation won’t work as well as in 2D domain. To address this issue, we propose a Query-Guided Prototype Adaption (QGPA) module to adapt the prototype from support point clouds feature space to query point clouds feature space. With such prototype adaption, we greatly alleviate the issue of large feature intra-class variation in point cloud and significantly improve the performance of few-shot 3D segmentation. Besides, to enhance the representation of prototypes, we introduce a Self-Reconstruction (SR) module that enables prototype to reconstruct the support mask as well as possible. Moreover, we further consider zero-shot 3D point cloud semantic segmentation where there is no support sample. To this end, we introduce category words as semantic information and propose a semantic-visual projection model to bridge the semantic and visual spaces. Our proposed method surpasses state-of-the-art algorithms by a considerable 7.90\% and 14.82\% under the 2-way~1-shot setting on S3DIS and ScanNet benchmarks, respectively. Code is available at \href{https://github.com/heshuting555/PAP-FZS3D}{https://github.com/heshuting555/PAP-FZS3D}.
\end{abstract}


%
\IEEEpeerreviewmaketitle

\section{Introduction}\label{sec:introduction}

\IEEEPARstart{P}{oint} cloud semantic segmentation aims to classify every point in a given 3D point cloud representation of a scene~\cite{landrieu2017large,qi2017pointnet}. It is one of the essential and fundamental tasks in the field of computer vision and is in intense demand for many real-world applications, \eg, virtual reality, self-driving vehicles, and robotics. Driven by the large-scale datasets and powerful deep learning technologies, fully supervised 3D point cloud semantic segmentation methods~\cite{huang2018recurrent,li2018pointcnn,li2022primitive3d,wang2019dynamic,li2023transformer} have demonstrated significant achievements in recent years. Nevertheless, it is laborious and expensive to build large-scale segmentation datasets with point-level annotations. Besides, additional annotated samples for novel categories and fine-tuning/re-training operations are required when extending the trained segmentation model to novel categories. To address these issues, few-shot 3D point cloud semantic segmentation is proposed~\cite{zhao2021few} and has attracted lots of attention. Few-shot point cloud segmentation aims to generate mask for the unlabeled point cloud of query sample based on the clues of a few labeled support samples. It greatly eases the heavy demand for large-scale datasets and demonstrates good generalization capability on novel categories.

\begin{figure}
\begin{center}
  \includegraphics[width = 0.486\textwidth]{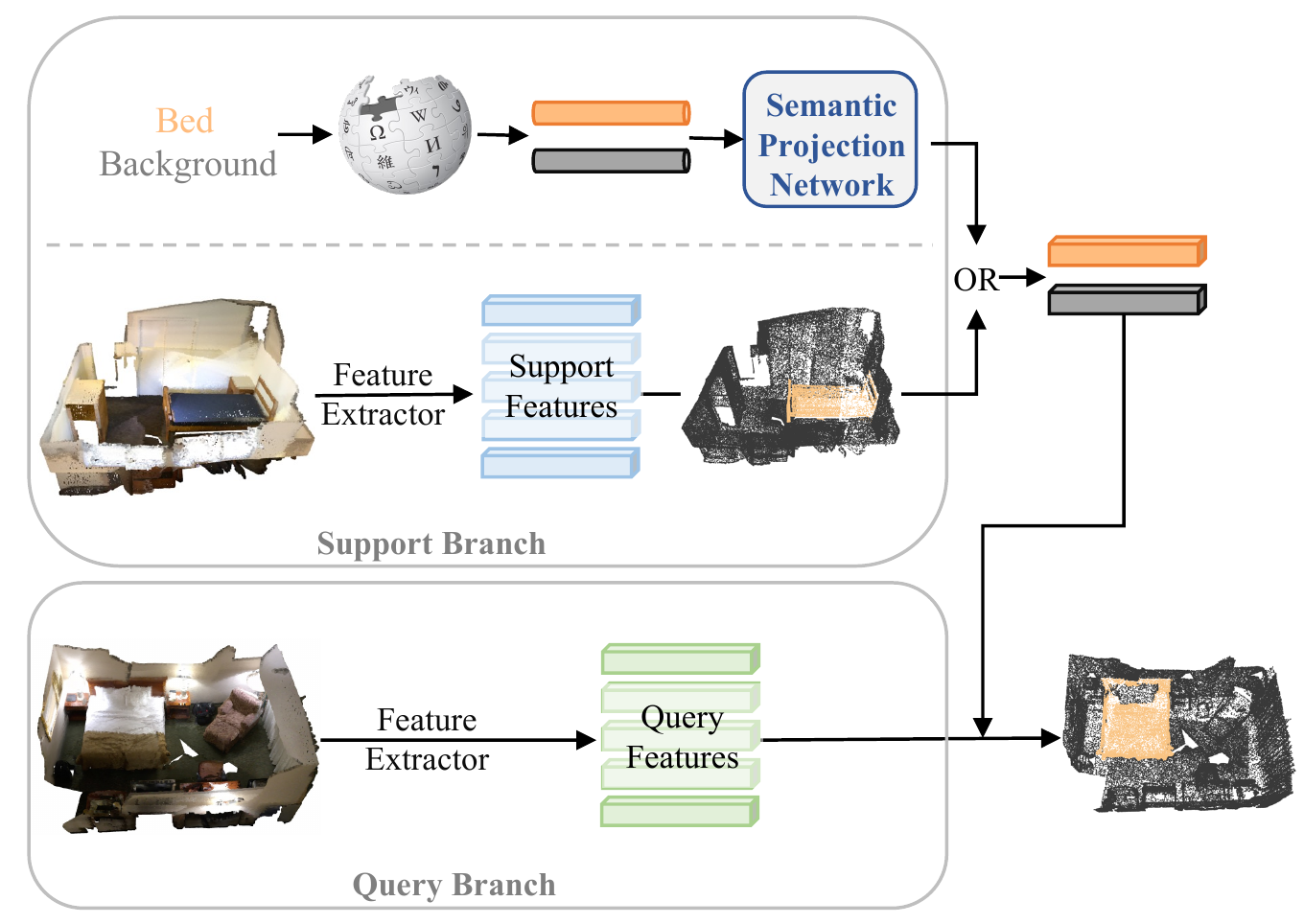}
\end{center}
\caption{Point-wise annotated masks are necessary for previous few-shot 3D point cloud segmentation methods. In this work, we introduce a semantic projection network, which generates prototypes for target categories via texts, as an alternative choice to make up for the limitations of the previous methods that cannot work without masks.}
\label{fig:1}
\end{figure}

A critical challenge of few-shot 3D point cloud segmentation lies in how to effectively classify every point by the limited support information.
Methods in 2D few-shot segmentation~\cite{PANet,liu2020part,liu2022few,yang2020prototype} extract discriminative and representative support features as prototypes (feature vectors) to guide the segmentation of query images, which has achieved significant results.
However, the success of few-shot semantic segmentation in 2D computer vision is driven by the pre-training on large-scale datasets like imagenet~\cite{deng2009imagenet}. The feature extractor pretrained on large-scale datasets greatly helps the few-shot learning by generating a good feature representation of objects. However, the development of 3D deep learning is hindered by the limited volume and instance modality of current datasets, due to the significant cost of 3D data collection and annotation. This results in less representative features and large feature variation in few-shot 3D point cloud segmentation, even for intra-class samples.
Therefore, the prototypical methods that work well in 2D few-shot segmentation are ineffective for the less-well pre-trained networks for 3D point cloud.
To address this issue, we propose a Query-Guided Prototype Adaptation (QGPA) module to modify features of support sample prototypes to  feature space of query sample. Specifically, the proposed QGPA leans the feature distribution mapping via cross attention between support and query features on the channel dimension, which produces a feature channel adaptation to convert the prototypes from support feature space to query feature space. Propagating the channel-wise distribution from support feature space to query feature space for prototypes smooths the channel distribution discrepancy. With such prototype adaptation, we greatly alleviate the feature variation issue in point clouds and significantly improve the performance of few-shot 3D point cloud semantic segmentation.

Moreover, optimizing prototype generation is worth enhancing the category-related features, as more representative and discriminative prototypes are the foundation for the success of subsequent adaptation and segmentation. If the prototype obtained from the support feature is not an apposite representative, it can hardly transfer informative clues to the query sample. 
Meantime, the usage of prototype adaptation reduces the influence of query mask supervision on prototype generation from the support set, despite the proposed QGPA having greatly narrowed down the feature gap between prototypes and query features.
Hence, we propose a Self-Reconstruction (SR) module, which enables prototypes to reconstruct the support masks, to strengthen the representation of prototypes. Specifically, after obtaining the prototype by an average pooling over the features of points indicated by the support mask, we apply the prototype back to the support features to reconstruct the support mask and employ explicit supervision on this self-reconstruction process. Such a simple self-reconstruction plays an important regularization role in the whole few-shot point cloud segmentation task to enhance the discriminative and semantic information embedded in prototypes.

Finally, although previous approaches~\cite{zhao2021few} and the above-proposed method reduce the number of required annotated samples via meta-learning, point-wise segmentation masks are still necessary. In some practical application scenarios, we may only have the category name of interest but have no corresponding images or masks.
Thus, in this work, we propose to step forward further and discard support masks, \ie, jointly considering few-shot and zero-shot 3D point cloud segmentation, as shown in Fig.~\ref{fig:1}. To this end, we introduce the semantic information, \eg, words of category name, to indicate the target categories and propose a Semantic Projection Network that bridges the semantic and visual features. Our projection network takes the semantic embedding as input and outputs a projected prototype, supervised by real prototypes from point clouds. 
During testing, besides obtaining prototypes via support branch with dense-annotated support masks, prototypes can be alternatively obtained by inputting semantic words with our proposed projection network.

In a nutshell, the main contributions of our work are:
\begin{itemize}
\setlength
  \item We propose an efficient and effective Query-Guided Prototype Adaption (QGPA) that propagates the channel-wise features from support sample prototypes to query feature space, which maps prototypes into query feature space. Prototypes are thus endowed with better adaption ability to mitigate channel-wise intra-class sample variation.
  \item We introduce Self-Reconstruction (SR) module that enforces the prototype to reconstruct the support mask generating this prototype, which greatly helps the prototype preserve discriminative class information.
  \item We design a semantic projection network to produce prototypes with the input of semantic words, which facilitates the inference without the use of support information.
  \item We achieve new state-of-the-art performance on two few-shot point cloud segmentation benchmarks, S3DIS and ScanNet. Specifically, our method significantly outperforms state-of-the-arts by 7.90\% and 14.82\% under the challenging 2-way-1-shot setting on S3DIS and ScanNet benchmarks, respectively.
\end{itemize}

\section{Related Work}
\subsection{3D Point Cloud Semantic Segmentation}
3D point cloud semantic segmentation aims to label each point in a given 3D point cloud by the most appropriate semantic category from a set of predefined categories. Thanks to the great success of deep neural networks, most recent deep-learning-based approaches have achieved impressive improvements in point cloud segmentation performance. There are two mainstreams in point cloud segmentation: voxel-based~\cite{graham20183d, SubmanifoldSparseConvNet, choy20194d} and point-based methods~\cite{qi2017pointnet, qi2017pointnet++, wu2019pointconv, zhao2020point, li2018pointcnn,lai2022stratified,vu2022softgroup}. The point-based methods have attracted more and more attention because of its simplicity and effectiveness. PointNet~\cite{qi2017pointnet}, a first point-based method, proposes a novel neural network to segment point clouds directly, which preserves the permutation invariance of the input well.
DGCNN \cite{wang2019dynamic} utilizes EdgeConv module to capture local structures which is neglected in PointNet. Despite these approaches achieved promising segmentation performance, they cannot easily segment unseen categories without being fine-tuned on enough labeled data. In this work, we follow the structure of DGCNN to capture local structure feature and propose our method to generalize to new classes with only a few of annotated samples.
 
 \subsection{Few-shot 3D Point Cloud Semantic Segmentation}
Few-shot 3D Point Cloud Semantic Segmentation put the general 3D point cloud semantic segmentation in a few-shot scenario, where model is endowed the ability to segment novel classes with only a handful support data.
Zhao~\etal~\cite{zhao2021few} propose attention-aware multi-prototype transductive inference to segment novel classes with a few annotated examples for few-shot point cloud semantic segmentation for the first time.
 However, attMPTI is very complicated and time-consuming due to exploiting multiple prototypes and establishing graph construction for few-shot point cloud segmentation and cannot achieve the impressive result.
 In this work, we deal with the few-shot point cloud semantic segmentation following the paradigm of~\cite{zhao2021few}. We explore to mitigate the feature variation for the objects with same label but from different images via a simple and effective transformer design.

\subsection{Few-shot Learning and Zero-shot Learning}
Few-shot learning focuses on learning a new paradigm for a novel task with only a few annotated samples.
Existing work can be grouped into two main categories, which are based respectively on metric learning~\cite{snell2017prototypical, vinyals2016matching, sung2018learning, zhang2020deepemd}, and meta-learning network~\cite{finn2017model, rusu2018meta, cai2018memory}.
The core concept in metric learning is distance measurement between images or regions.
For example, 
Vinyals~\etal~\cite{vinyals2016matching} design matching networks to embed image into an embedded feature and implement a weighted nearest neighbor matching for classifying unlabelled samples. 
Snell~\etal~\cite{snell2017prototypical} introduce a prototypical network to build a metric space where an input is identified in line with its distance from the class prototypes. 
Our work is in conformity with the prototypical network while we use it for more challenging segmentation tasks with a simple yet effective design.

Zero-shot learning~\cite{lampert2009learning,D2Zero,zhang2021prototypical,PADing} aims to classify images of unseen categories with no training samples via utilizing semantic descriptors as auxiliary information. There are two main paradigms: classifier-based methods and instance-based methods. Classifier-based methods aim to learn a good projection between visual and semantic spaces~\cite{demirel2017attributes2classname,li2018deep} or transfer the relationships obtained in semantic space to visual feature space~\cite{gan2015exploring,zhang2016zero}. Another main branch is instance-based methods~\cite{yu2013designing} that synthesize some fake samples for unseen classes. The proposed semantic projection network bridges semantic prototypes and visual prototypes, and combines zero-shot learning with few-shot learning to flexibly handle cases with masks and without masks, which greatly eases the heavy demand for large-scale 3D datasets.

\subsection{Few-shot Segmentation}

As an extension of few-shot classification, few-shot segmentation performs a challenging task of predicting dense pixel classification~\cite{ding2018context,shuai2018toward,SVCNet,BFP} on new classes with only a handful of support examples~\cite{ding2023self,liu2021few}. 
Shaban~\etal~\cite{OSLSM} introduce this research problem for the first time and design a classical two-branch network following the pipeline of Siamese network \cite{koch2015siamese}.
Later on, PL~\cite{dong2018few} introduces the concept of prototypical learning into segmentation task for the first time.
where predictions is generated according to the cosine similarity between pixels in query image and prototypes generated from support image and support mask.
SG-One~\cite{SG-one} designs masked average pooling to generate object related corresponding prototype, which has become the cornerstone of subsequent methods.
PANet~\cite{PANet} extends this work to a more efficient design and propose a prototype alignment regularization to make better use of support set information, achieving better generalization capability.
CANet~\cite{zhang2019canet} designs two-branch architecture to perform multi-level feature comparison and embed iterative optimization module to get refined predicted results.
PPNet~\cite{liu2020part} and PMMs~\cite{yang2020prototype} have similar idea to decompose objects into multiple parts and are capable of obtaining diverse and fine-grained representative features.
PFENet~\cite{tian2020pfenet} generates training-free prior masks utilizing a pre-trained backbone, and alleviated the spatial inconsistency through enhancing query features with prior masks and support features.

However, existing methods have a common characteristic, \ie the feature extractor pretrained on large-scale datasets greatly affects the performance of few-shot learning. Therefore, the feature extractor for few-shot 3D point cloud cannot provide representative features for objects because of lacking of pretraining on large-scale datasets hindered by the limiting volume and instance modality. Consequently, existing popular prototypical methods in 2D few-shot classification/segmentation do not work well in the field of 3D point cloud segmentation. In this work, we tackle this issue by proposing a prototype adapter and self-reconstruction to project the prototype
from support point clouds feature space to query point clouds feature space effectively.

\section{Approach}
In this section, we present the proposed approach. We first give the task definition in Sec.~\ref{sec:defination} and the architecture overview of our proposed model in Sec.~\ref{sec:architecture}. Then the proposed Query-Guided Prototype Adaption (QGPA), Self-Reconstruction, and Semantic Prototype Projection are presented in Sec.~\ref{sec:QGPA}, Sec.~\ref{sec:SPP}, and Sec.~\ref{sec:SR}, respectively.

\begin{figure*}[t]
\begin{center}
  \includegraphics[width = 0.956\textwidth]{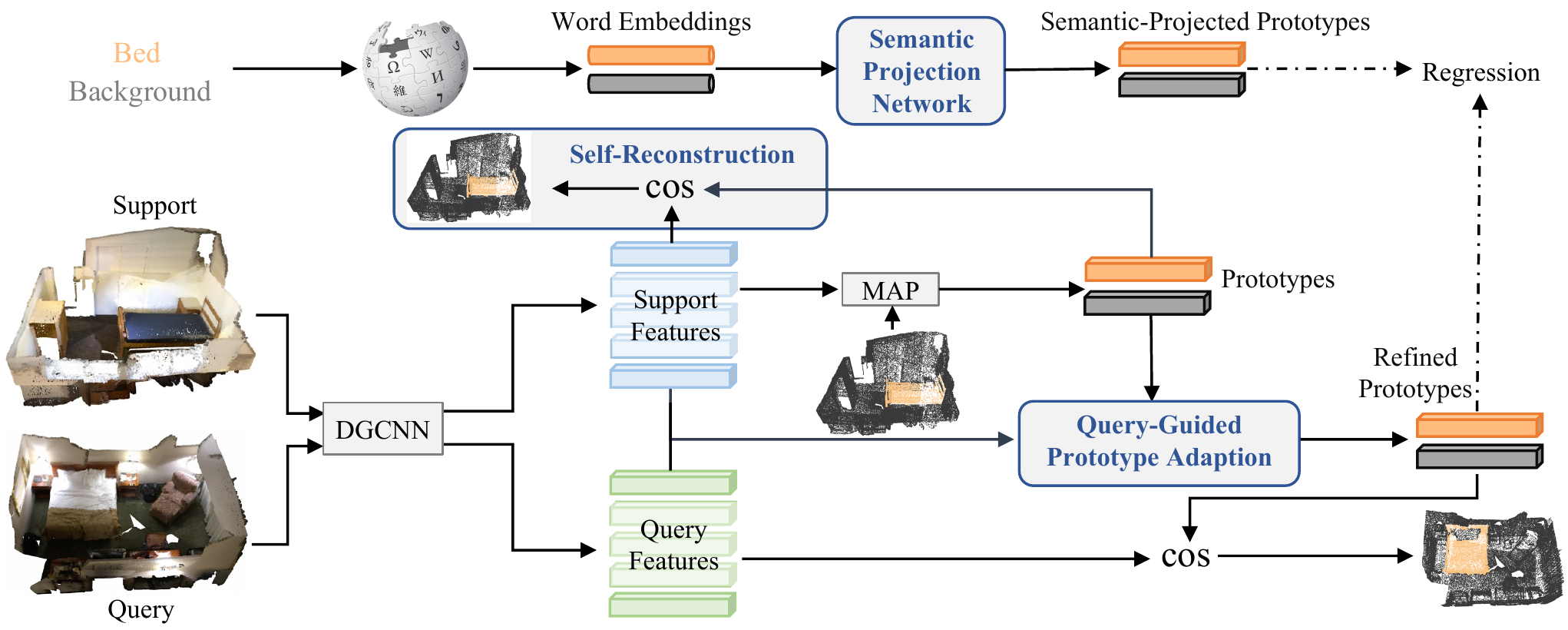}
\caption{Architecture overview of our training model. We embed the support and query point clouds into deep features by DGCNN with shared weights. The prototypes are generated by masked average pooling (MAP) over the support features. 
We further introduce Query-Guided Prototype Adaption (QGPA) and Self-Reconstruction (SR) to enhance the discriminative and representative of prototypes. A semantic projection network is proposed to replace visual prototypes with semantic prototypes to get rid of support branch during inference. The query images are segmented by computing the pixel-wise similarity between the adapted prototypes and feature maps. ``cos'' denotes cosine similarity.}
\label{Fig:framework}
\end{center}
\end{figure*}

\subsection{Problem Definition}\label{sec:defination}

We follow previous work~\cite{zhao2021few} to define the training and testing protocols for few-shot point cloud semantic segmentation. Each point cloud $I$ $\in \mathbb{R}^{N\times(3+f_0)}$ contains $N$ points associated with the coordinate information $\in \mathbb{R}^3$ and an additional feature $\in \mathbb{R}^{f_0}$, \eg, color. Suppose we have point cloud samples from two non-intersection sets of classes $\mathcal{C}_{\text{seen}}$ and $\mathcal{C}_{\text{unseen}}$. We train our model on a train set $\mathcal{D}_{\text{train}}$ that is constructed from $\mathcal{C}_{\text{seen}}$ and evaluate on the test set $\mathcal{D}_{\text{test}}$ 
built from $\mathcal{C}_{\text{unseen}}$. Widely used episodic paradigm~\cite{vinyals2016matching} in few-shot learning is adopted to construct the train set $\mathcal{D}_{\text{train}}$ and test set $\mathcal{D}_{\text{test}}$. 

Each training/testing episode in few-shot point cloud segmentation instantiates a $C$-way $K$-shot segmentation learning task. There are $K$ $\langle$point cloud, mask$\rangle$ pairs for each of the $C$ different categories in support set, and every point in the query point cloud is classified to one of the $C$ categories or ``background'' that does to belong to any of the $C$ categories. We denote the support set as $\mathcal{S}=\{(I_S^{c,k},M_S^{c,k})\}$, where $I$ is point cloud, $M$ is mask, $k\in\{1,\cdots,K\}$, and $c\in\{1,\cdots,C\}$. At each episode, the inputs are $\{I_S^{c,k}, M_S^{c,k}, I_Q\}$ and the output is segmentation prediction of query point cloud $I_Q$, of which the ground truth is $M_Q$. During training, the model gains knowledge about the $C\in C_{train}$ classes from the support set and then applies the knowledge to segment the query set. After obtaining the segmentation model from the training set $D_{train}$, the model's few-shot segmentation performance is evaluated on the testing set $D_{test}$. As each training episode contains different semantic categories, the model can be well generalized after training. 

We further introduce semantic words as an alternative choice to provide support information for target categories. In our training stage, the support set are reformulated as $\mathcal{S}=\{(I_S^{c,k}, M_S^{c,k}, W_S)\}$, where $W_S$ is semantic word set for support point cloud. In the testing stage, different from previous methods that compulsorily require point-level annotated masks, it is acceptable for our approach to only input the semantic word related to the target-of-interest, \ie, $\mathcal{S}=\{W_S^{c}\}$. The goal of our approach is summarized as, to train a model that, when given support set $S$ with either annotated masks or semantic words that provide support information for $C$ classes, generates a segmentation mask with point-wise labels from the supported $C$ classes (and ``background'') for the input query point cloud $Q$.

\subsection{Architecture Overview}\label{sec:architecture}

The overall training architecture of our proposed approach is shown in \figurename~\ref{Fig:framework}. For each episode in the training stage, point clouds of the support set and query set are processed by a DGCNN~\cite{wang2019dynamic} backbone and mapped to deep features. To obtain prototypes from support set, the masked average pooling (MAP) is applied over support features. Then, Query-Guided Prototype Adaption (QGPA) is utilized to rectify the feature channel distribution gap between query and support point clouds. The cosine similarity is employed between prototype and query feature to produce the score maps for generating the final mask prediction. Every point cloud in the query set is assigned the label with the most similar prototype. To preserve class-related discriminative information embedded in prototype, Self-Reconstruction (SR) module is introduced to obtain self-consistency high-quality prototypes. What's more, a semantic projection network is proposed to project the semantic word embeddings to visual prototypes under regression loss. During the inference stage, the proposed semantic projection network can take place of the visual support branch to provide prototypes when there is no point-wise annotated masks as support information.

\subsection{Query-Guided Prototype Adaption}\label{sec:QGPA}
We generate prototypes for every target category in the support set via conducting masked average pooling over the support features with corresponding support masks. Given a support set $S=\{(I_S^{c,k},M_S^{c,k})\}$, where $k\in\{1,...,K\}$ and $c\in\{1,...,C\}$ are the C-way and K-shot indexes respectively, and its feature $F_S^{c,k}\in \mathbb{R}^{N \times d}$, where $N$ is the number of points and $d$ is feature channel number, the prototype (feature vector) of category $c$ is obtained by:
\begin{equation}
    p^c=\frac{1}{K}\sum_k\frac{\sum_{x}F^{c,k}_{S,x}\mathbbm{1}(M^{c,k}_{S,x}=c)}{\sum_{x}\mathbbm{1}(M^{c,k}_{S,x}=c)},
\label{Eq:1}
\end{equation}
where $x\in\{1,...,N\}$ denotes the coordinate positions and $\mathbbm{1}(*)$ is the binary label indicator that outputs $1$ when $*$ is true. Besides the target categories, we compute a background prototype $p^{0}$ to represent the points that do not belong to any of the $C$ target categories:
\begin{equation}
    p^{0}=\frac{1}{CK}\sum_{c,k}\frac{\sum_{x}F^{c,k}_{S,x}\mathbbm{1}(M^{c,k}_{S,x}\notin \{1,...,C\})}{\sum_{x}\mathbbm{1}(M^{c,k}_{S,x}\notin\{1,...,C\})}.
\label{Eq:2}
\end{equation}
Now we have a set of prototypes $\mathbbm{P}=\{p^0, p^1,...,p^C\}$.

\begin{figure}[t]
	\centering
	\includegraphics[width = 0.42\textwidth]{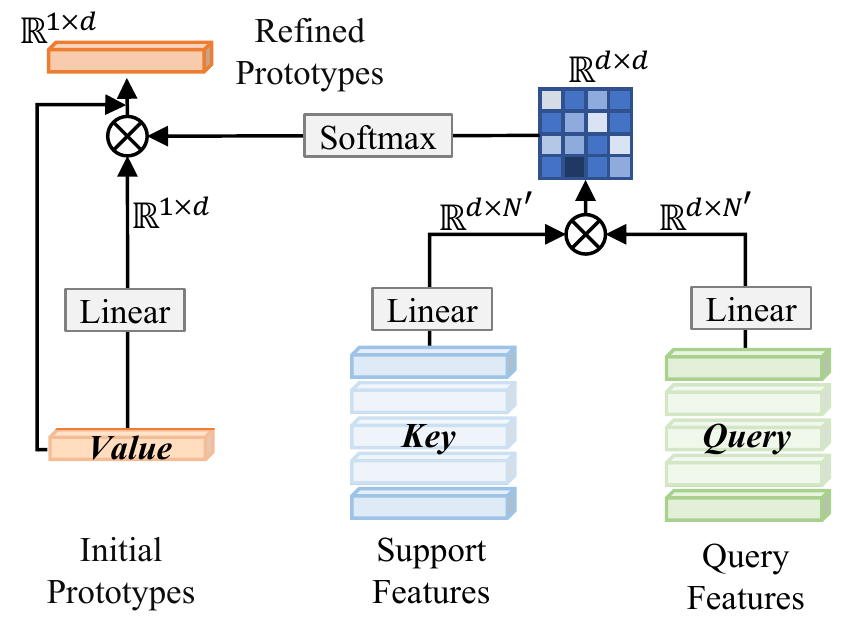}
	\caption{\small{The architecture of our proposed Query-Guided Prototype Adaption (QGPA) module. Query Features and Support Features are extracted from the Query point clouds and Support point clouds by DGCNN~\cite{wang2019dynamic} (see \figurename~\ref{Fig:framework}).}}
	\label{fig:cross_attention}
\end{figure}

Prototypes obtained from the support point clouds have a channel-wise feature distribution gap with features of query point clouds, as we discussed in Sec.~\ref{sec:introduction}. Each sample has different feature channel response distribution~\cite{chu2016structured}. This feature distribution gap is more obvious in 3D point cloud than in 2D segmentation in terms of image level, due to the lack of large-scale 3D datasets for pretraining of feature extractor. To rectify the feature distribution gap, we design Query-Guided Prototype Adaption~(QGPA) that maps the prototype to query feature space under the guidance of query-support feature interaction, as shown in \figurename~\ref{fig:cross_attention}. 

In detail, given a prototype $p^i\in \mathbb{R}^{1 \times d}$, $i\in\{0, 1,..., C\}$, its support features that generate this prototype, and a query feature $F_Q\in \mathbb{R}^{N \times d}$, we calculate the channel-wise cross attention between support and query features, producing a projection attention matrix. The prototype is then processed by the projection attention matrix to fit the query feature distribution. We first average these support features for each prototype:
\begin{equation}%
F_S^i =\left\{
\begin{aligned}
&\frac{1}{K}\sum_k F_S^{c,k},&i&\in\{1,...,C\}, \\
&\frac{1}{CK}\sum_{c,k}F_S^{c,k},&i&=0,
\end{aligned}
\right.
\label{eq:3}
\end{equation}
where $F_S^i \in \mathbb{R}^{N \times d}$ represent averaged support feature for prototype $p^i$. 
Then the input to our QGPA is designed as:
\begin{align}
    {Query} = {F_Q}^\top \textbf{W}_q, \;\;  
    {Key} = {F_S^i}^\top \textbf{W}_k, \;\;
    {Value} = p^i\textbf{W}_v,
    \label{eq:qkv}
\end{align}
where $\textbf{W}_q$,$\textbf{W}_k$ $\in \mathbb{R}^{N\times N'}$ and $\textbf{W}_v  \in \mathbb{R}^{d\times d}$
are the learnable parameters for fully connected layers that project features and prototype to the corresponding latent space, respectively. $N'\leq N$ is decreased hidden point number to save computing resources. Hence, we get $Query$ $\in \mathbb{R}^{d\times N'}$, $Key$ $\in \mathbb{R}^{d\times N'}$ and $Value$ $\in \mathbb{R}^{1\times d}$ for transformer attention.

Then, we calculate a matrix multiplication between the $Query$ and the transpose of $Key$, and add a softmax layer to obtain the channel attention map $Attn \in \mathbb{R}^{d\times d}$
as:
\begin{equation}
    Attn = \operatorname{softmax}(\frac{{Query} \cdot{Key}^\top}{\sqrt{d}}),
    \label{eq:attn}
\end{equation}
where $\operatorname{softmax}(\cdot)$ is a row-wise softmax function for attention normalization.
This cross-attention map \textit{Attn} establishes the channel-to-channel correspondence between query and support features, which guides the channel distribution propagation.
Finally, a matrix multiplication is conducted between the \textit{Attn} and the transpose of value to rectify the prototype to query feature channel distribution:
\begin{equation}
    \dot{p}^i = p^i + 
    \textbf{W}_p (Attn \cdot{Value}^\top)^\top,
    \label{eq:get_P}
\end{equation}
where the original prototype is added as a residual connection for more stable training convergence, $\textbf{W}_p\in \mathbb{R}^{d\times d}$ is the parameter of a fully connected layer. All the $C+1$ prototypes $\mathbbm{P}=\{p^0, p^1,...,p^C\}$ are dealt with Eq.~(\ref{eq:3}) to Eq.~(\ref{eq:get_P}), generating $C+1$ refined prototypes denoted as $\dot{\mathbbm{P}}=\{\dot{p}^0, \dot{p}^1,...,\dot{p}^C\}$. With Eq.~(\ref{eq:get_P}), the prototype generated by the support set is in line with the feature channel distribution of the query sample, which contributes to mitigating feature variation between point clouds sample. Therefore, when we perform cosine similarity between query features and refined prototypes, channel-wise inconsistency will be alleviated, which greatly 
contributes to getting a better result.

We employ cosine similarity with softmax function between query feature $F_Q$ and prototypes $\dot{\mathbbm{P}}$ to get probability score map. For each prototype $\dot{p}^i$ we have the score map:
\begin{equation}
S^{i}_{Q, x}\!=\!\frac{exp(-{\alpha}\langle F_{Q,x}, \dot{p}^i\rangle)}{\sum_{\dot{p}^i\in\dot{\mathbbm{P}}}exp(-{\alpha}\langle F_{Q,x}, \dot{p}^i\rangle)},
\end{equation}
where $\langle a, b\rangle$ represents the computation of cosine similarity between $a$ and $b$, $\alpha$ is an amplification factor. 
The predicted segmentation mask is then given by 
\begin{equation}
    \hat{M}_{Q, x}=\mathop{\arg\max}_{i}S^{i}_{Q, x}.
\label{eq:predictedmask}
\end{equation}
Learning proceeds by minimizing the negative log-probability 
\begin{equation}
    \mathcal{L}_{seg}=-\frac{1}{N}\sum_{x}\sum_{i}\mathbbm{1}({M}_{Q,x}=i)logS^{i}_{Q, x},
\label{eq:loss_seg}
\end{equation}
where $N$ is the total number of points, and ${M}_{Q}$ is the ground truth mask of query point cloud ${I}_{Q}$.

\subsection{Self-Reconstruction}\label{sec:SR}

Although through section \ref{sec:QGPA} we can obtain refined prototypes that better fit the distribution of query features, 
they may lose their original critical class and semantic information that were learned from the support set. Additionally, it’s crucial to extract more representative and discriminative prototypes from the support set, as this is the foundation for the success of subsequent adaptation and segmentation.
The discriminative prototypes should have the category information flow of the support point cloud, \ie, prototypes need to have the capability to reconstruct ground-truth masks from themselves.

For each support feature $F_{S}^{c,k}$ and corresponding support mask $M^{c,k}_{S}$, we calculate cosine similarity with softmax function between support feature $F_{S}^{c,k}$ and prototypes $\{p^0,p^c\}$, to get score map $S_{S}^{c,k}$,
\begin{equation}
S^{c,k,i}_{S, x}\!=\!\frac{exp(-{\alpha}\langle F_{S,x}^{c,k}, {p}^i\rangle)}{\sum_{{p}^i\in{\{p^0,p^c\}}}exp(-{\alpha}\langle F_{Q,x}, {p}^i\rangle)}.
\end{equation}
The reconstructed support mask is given by
\begin{equation}
    \hat{M}_{S,x}^{c,k}=\mathop{\arg\max}_{i}S^{c,k,i}_{S, x}.
\label{eq:rmask}
\end{equation}
This reconstructed support mask $\hat{M}_{S}^{c,k}$ is expected to be consistent with the information of the original support point cloud ground truth mask $M^{c,k}_{S}$. We call this process as Self-Reconstruction~(SR). The Self-Reconstruction loss $\mathcal{L}_{\text{sr}}$ is computed by minimizing the negative log-probability, similar to Eq.~(\ref{eq:loss_seg}):
\begin{equation}
\label{eqn:loss_sr}
    \mathcal{L}_{\text{sr}} = -\cfrac{1}{CKN}\sum_{c,k,x} \sum_{i}\mathds{1}(M_{S,x}^{c,k}=i)\log  S^{c,k,i}_{S, x}.
\end{equation}
The final segmentation loss is sum of $\mathcal{L}_{seg}$ and $\mathcal{L}_{sr}$:
\begin{equation}
 \mathcal{L}_{total} =  \mathcal{L}_{seg} + \mathcal{L}_{sr}.
\label{eq:final_loss}
\end{equation}

Without this Self-Reconstruction as a constraint, prototypes may lose the original critical class and semantic information when aligning with the distribution of the query feature. On the other hand, it does not adequately utilize the
support information for few-shot learning. The proposed Self-Reconstruction module serves as an important regularization role in the whole few-shot point cloud segmentation task to preserve discriminative and semantic information embedded in prototypes. 
Besides, it provides a mechanism to balance prototype adaptation while maintaining the original separability, adding an additional level of refinement to the process.

\subsection{Semantic Prototype Projection}\label{sec:SPP}

Few-shot learning methods though reduce the number of required annotated samples to a certain degree, point-level segmentation masks are still compulsory during inference, as in Eq.~(\ref{Eq:1}) and in Eq.~(\ref{Eq:2}). To relieve the demanding requirements for point-level annotations, we step forward further and explore discarding support masks in this work. To this end, we introduce the semantic information to indicate the target categories and propose a semantic projection network that projects the semantic words from semantic space to visual space, \ie, from semantic words to visual prototypes. Given a set of semantic words $W_S=\{w_S^0,w_S^1,...,w_S^C\}$, where $w_S^i$ represents the name word for class $i$, we first employ a text-encoder, \eg, word2vec~\cite{word2vector} or CLIP~\cite{clip}, to produce corresponding semantic embeddings, denoted as $E_S=\{e_S^0,e_S^1,...,e_S^C\}$, so that we have $e_S^i=textencoder(w_S^i)$. A dump of the Wikipedia corpus containing 3 billion words is utilized to train the word2vec model and 400 million text-image pairs are used in training CLIP.
Such text encoder has the ability to produce corresponding embeddings for words that have the corresponding relationships in context~\cite{ding2021vision,ding2020phraseclick,VLTPAMI}. Hence, the semantic embeddings from word2vec/CLIP is expected to have obtained the semantic relationship among the classes.
We utilize \texttt{Linear+LeakyReLU+Dropout} to form a projection block, whose output is produced by another \texttt{Linear} transformation to generate the projected prototype. We denote the projected prototypes as $\mathbbm{\hat{P}}=\{\hat{p}^0, \hat{p}^1,...,\hat{p}^C\}$, then we have $\hat{p}^i=Projection(e_S^i)$. We follow~\cite{GMMN} to use a differential criterion to compare the target prototypes $\dot{\mathbbm{P}}$ and the projected prototypes $\mathbbm{\hat{P}}$ by minimize the maximum mean discrepancy:
 \begin{equation}
    \mathcal{L}_{reg}=\sum_{\dot{p},\dot{p}'\in\dot{\mathbbm{P}}}\mathcal{G}(\dot{p},\dot{p}')+\sum_{\hat{p},\hat{p}'\in\hat{\mathbbm{P}}}\mathcal{G}(\hat{p},\hat{p}')-2\sum_{\dot{p}\in\dot{\mathbbm{P}}}\sum_{\hat{p}\in\hat{\mathbbm{P}}}\mathcal{G}(\dot{p},\hat{p}),
\label{Eq:gmmn}
\end{equation}
where $\mathcal{G}(a,b)=exp(-\frac{1}{2\sigma^2}\parallel\!a-b\!\parallel^2)$ is a Gaussian function with bandwidth parameter $\sigma$. The notation $\dot{p}'$ represents one of the points in the sample $\dot{\mathbbm{P}}$ that is being compared to the point $\dot{p}$ via the kernel function. 
When the training is over, the mapping from semantic embedding to visual prototype has been established. During the testing process, the proposed projection network can take place of the support branch to provide prototypes when no point-level annotated masks are provided. As long as the name of the category of interest is given, the corresponding prototype with semantic features is generated, and then query point cloud can be segmented without point-wise annotated masks as support information.

\section{Experiments}
In this section, we report the experimental results of the proposed approach in comparison with previous state-of-the-art methods and ablation studies that verify the effectiveness of our proposed modules.
\subsection{Implementation Details}

Our approach is implemented based on the public platform Pytorch\footnote{http://pytorch.org}. 
During training, we employ a variety of techniques to augment the training samples, such as Gaussian jittering, random shift, random scale, and random rotation around the z-axis. The training samples are then randomly sampled at every episode of each iteration.
The feature extractor is initially pre-trained on the training set $\mathcal{D}_{\text{train}}$ for a total of 100 epochs, using Adam as the optimizer. We set the learning rate and batch size to 0.001 and 32, respectively. Following this pre-training stage, we proceed to train our proposed model with the pre-trained weights as initialization weights. 
For the convenience of explanation, we divide the network into two sub-networks, the segmentation part (including feature extractor, Query-Guided Prototype Adaption, and Self-Reconstruction) and the semantic projection part (Semantic Projection Network).
To ensure that the performance of segmentation part is not affected by the semantic projection part, we employ two independent optimizers during joint training.
Adam is utilized as the optimizer to train the segmentation part, with an initial learning rate of 0.001 for the newly added layers, and 0.0001 for the feature extractor. 
It should be noted that the Self-Reconstruction does not introduce any additional parameters. 
The learning rates decay by 0.5 after every 5K iterations.
The semantic projection component is trained using the Adam optimizer with a consistent learning rate of 0.0002 throughout the entire training process. 
As for hyper-parameter settings, we set the $\sigma$ of $\mathcal{G}$ used in Eq.~(14) to $\{2,5,10,20,40,60\}$.
We set the number of Transformer layers and attention head to 1 for simplicity. 

\subsection{Datasets and Evaluation Metrics}
\paragraph{Datasets}
We perform experimental evaluation on two public 3D point clouds datasets \textbf{S3DIS} \cite{armeni20163d} and \textbf{ScanNet} \cite{dai2017scannet}.
S3DIS is composed of 271 point clouds from Matterport scanners in six different areas from three buildings. 
The annotation for the point clouds has 12 semantic classes in addition to one background class annotated with clutter.
ScanNet is made up of 1,513 point clouds of scans from 707 unique indoor scenes. 
The annotation for the point clouds has 20 semantic classes in addition to one background class annotated with unannotated space.
Since the original scene is too large to process, we need to split it into smaller blocks.
As a result, S3DIS and ScanNet contain 7,547 and 36,350 blocks through data pre-processing strategy utilized in \cite{wang2019dynamic, qi2017pointnet}, respectively. $N=2,048$ points are randomly sampled for each block and each point is represented by a 9D vector (XYZ, RGB and normalized spatial coordinates). 

\begin{table*}[htbp]
\small
	\begin{center}
	\setlength\tabcolsep{8pt}
	\renewcommand\arraystretch{1.3}
	\scalebox{0.996}{
		\begin{tabular}{ C{0.6cm}  C{0.86cm}?C{0.6cm} C{0.6cm} C{0.6cm}? C{0.6cm} C{0.6cm} C{0.6cm} ? C{0.6cm}  C{0.6cm}  C{0.6cm} ? C{0.6cm} C{0.6cm} C{0.6cm}}

			\specialrule{.1em}{.05em}{.05em}
			\multirow{3}{*}{QGPA}&\multirow{3}{*}{SR}
			& \multicolumn{6}{c?}{\textbf{S3DIS}}
			& \multicolumn{6}{c}{\textbf{ScanNet}} \\ 
			\cline{3-14} 
			&& \multicolumn{3}{c?}{\textbf{1-shot}} & \multicolumn{3}{c?}{\textbf{5-shot}}
			& \multicolumn{3}{c?}{\textbf{1-shot}} & \multicolumn{3}{c}{\textbf{5-shot}}
			\\ \cline{3-14} 
			&& \multicolumn{1}{c}{S$^0$} & \multicolumn{1}{c}{S$^1$} &\multicolumn{1}{c?}{mean} 
			& \multicolumn{1}{c}{S$^0$} & \multicolumn{1}{c}{S$^1$} 
			& \multicolumn{1}{c?}{mean}              
			& \multicolumn{1}{c}{S$^0$} & \multicolumn{1}{c}{S$^1$} &\multicolumn{1}{c?}{mean} 
			& \multicolumn{1}{c}{S$^0$} & \multicolumn{1}{c}{S$^1$} 
			& \multicolumn{1}{c}{mean}  
			\\  \hline
		\xmarkg& \xmarkg &52.17  &57.85   &55.01      & 63.82 &67.60   &65.71      &41.03 &41.36   &41.19    &54.98   &51.10 &53.04 \\
		\cmark& \xmarkg &56.77   &61.19    &58.98     & 64.57 &69.16   &66.87      &52.66 &51.05   &51.85    &62.19   &57.23 &59.71 \\
		\xmarkg& \cmark &52.42  &57.94    &55.18     & 62.54 &69.36   &65.95      &41.02 &41.58   &41.30    &55.57   &51.13 &53.35 \\
		 \cmark& \cmark &59.45   &66.08     &62.76    & 65.40 &70.30   &67.85      &57.08 &55.94   &56.51    &64.55   &59.64 &62.10 \\
			\specialrule{.1em}{.05em}{.05em}
		\end{tabular}
	}
	\vspace{-3mm}
	\caption{\small{Results under \textbf{2-way}
setting on \textbf{S3DIS} and \textbf{ScanNet} dataset using mean-IoU metric (\%). QGPA and SR are our proposed Quey-Guided Prototype Adaption and Self-Reconstruction, respectively. S$^i$ denotes the split $i$ is used for testing.}\label{ablation}}
\end{center}
\end{table*}

Following~\cite{zhao2021few}, semantic classes in each dataset is evenly split into two non-overlapping subsets .
For both S3DIS and ScanNet, when testing the model with test class set $C_{unseen}$
on one fold, we use the other fold to train the model with train class set $C_{seen}$
for cross-validation.

In training process, an episode is constructed using the following procedure. First, $C$ classes from $C_{seen}$ is randomly chosen which should meet the criterion $N < |C_{seen}|$; Next, random choose sample from support set $S$ and a query set $Q$ based on the chosen $C$ classes. 
Finally, the ground-truth mask ${M_S}$ for the support set and ${M_Q}$ for the query set are generated from the original mask annotation as the binary mask according to the chosen classes. 
The episodes for testing are built in a similar form. Except for one difference, we traverse $N$ classes out of $C_{unseen}$ classes instead of randomly choosing $N$ classes to get more fair result. 100 episodes are sampled for evaluation.

\paragraph{Evaluation Metrics}
Following conventions in the point cloud semantic segmentation community, we evaluate all methods with Mean Intersection-over-Union (mean-IoU). The per-class Intersection over Union (IoU) is defined as $\frac{TP}{TP+FN+FP}$, where the $TP$, $FN$, and $FP$ is the count of true positives, false negatives and false positives, respectively. 
For few-shot setting, mean-IoU is calculated by averaging over all the classes in testing classes $\mathcal{C}_{\text{unseen}}$.

\begin{table}[htbp]
	\centering
	\renewcommand\arraystretch{1.2}
	\scalebox{0.9}{
		\begin{tabular}{C{1.5cm} C{1.5cm} C{1.5cm} | C{1.5cm} ? C{1.5cm} }
			\specialrule{.1em}{.05em}{.05em}
			AUG & MS & AL & S3DIS & ScanNet \\ \hline 
		    \xmarkg  & \xmarkg  & \xmarkg &  48.39 & 33.92    \\
			\cmark  & \xmarkg  & \xmarkg &  49.20 & 36.12    \\
			\cmark  & \cmark  & \xmarkg &  51.25 & 39.45  \\
			\cmark  & \cmark  & \cmark &  52.17 & 41.03   \\
			\specialrule{.1em}{.05em}{.05em}
	\end{tabular}}
	\caption{\small{Effects of different baseline network configuration under \textbf{2-way 1-shot} setting on S3DIS (S$^0$) and ScanNet (S$^0$) datasets.The symbols \cmark and \xmark indicate that the corresponding setting is included or excluded, respectively. The abbreviations AUG, MS, AL denote augmentation including random scale and random shift augmentations, multi-scale feature, align loss\cite{PANet},
respectively.}}
	\label{tab:ablation-baseline}
\end{table}

\begin{table*}[htbp]
\small
	\begin{center}
	\renewcommand\arraystretch{1.3}
	\scalebox{0.996}{
		\begin{tabular}{ L{1.6cm}? C{1.0cm}? C{1.0cm}?C{0.6cm} C{0.6cm} C{0.6cm}? C{0.6cm} C{0.6cm} C{0.6cm} ? C{0.6cm}  C{0.6cm}  C{0.6cm} ? C{0.6cm} C{0.6cm} C{0.6cm}}

			\specialrule{.1em}{.05em}{.05em}

			\multirow{3}{*}{Method}&\multirow{3}{*}{\#Params.}&\multirow{3}{*}{FLOPs}& \multicolumn{6}{c?}{\textbf{S3DIS}}
			& \multicolumn{6}{c}{\textbf{ScanNet}} \\ 
			\cline{4-15} 
			&&& \multicolumn{3}{c?}{\textbf{1-shot}} & \multicolumn{3}{c?}{\textbf{5-shot}}
			& \multicolumn{3}{c?}{\textbf{1-shot}} & \multicolumn{3}{c}{\textbf{5-shot}}
			\\ \cline{4-15} 
			&  &  & \multicolumn{1}{c}{S$^0$} & \multicolumn{1}{c}{S$^1$} &\multicolumn{1}{c?}{mean} 
			& \multicolumn{1}{c}{S$^0$} & \multicolumn{1}{c}{S$^1$} 
			& \multicolumn{1}{c?}{mean}              
			& \multicolumn{1}{c}{S$^0$} & \multicolumn{1}{c}{S$^1$} &\multicolumn{1}{c?}{mean} 
			& \multicolumn{1}{c}{S$^0$} & \multicolumn{1}{c}{S$^1$} 
			& \multicolumn{1}{c}{mean}  
			\\  \hline
		Baseline   &352.19K &7.12G    &52.17   &57.85    &55.01  &63.82   &67.60  &65.71      &41.03 &41.36   &41.19    &54.98   &51.10 &53.04  \\
		CWT\cite{CWT} &685.74K&8.16G  &52.14   &57.86    &55.00  &61.64   &66.48  &64.06  &42.33   &41.78   &42.05   &55.60   &53.77   &54.68 \\
		DETR\cite{detr} &2.62M  &8.20G &50.87   &54.09    &52.48  &57.29   &65.31  &61.30   &43.32   &45.07   &44.19   &55.42   &53.49   &54.45 \\
		D-QGPA      & 2.48M &7.48G    &51.55   &54.05    &52.80  &58.45   &65.97  &62.21  &36.08   &39.26   &37.67   &54.07   &51.19   &52.63   \\
		QGPA & 2.48M &7.48G &56.77   &61.19    &58.98     & 64.57 &69.16   &66.87 &52.66 &51.05   &51.85    &62.19   &57.23 &59.71  \\
			\specialrule{.1em}{.05em}{.05em}
		\end{tabular}
	}
	\vspace{-2mm}
	\caption{\small{Comparison of different transformer-like modules under \textbf{2-way 1-shot}
setting on \textbf{S3DIS} and \textbf{ScanNet} dataset using mean-IoU metric (\%). S$^i$ denotes the split $i$ is used for testing.}}
\label{tab:transformer-variations}
\end{center}
\end{table*}

\begin{figure*}[htbp]
	\centering
	\includegraphics[scale=0.53]{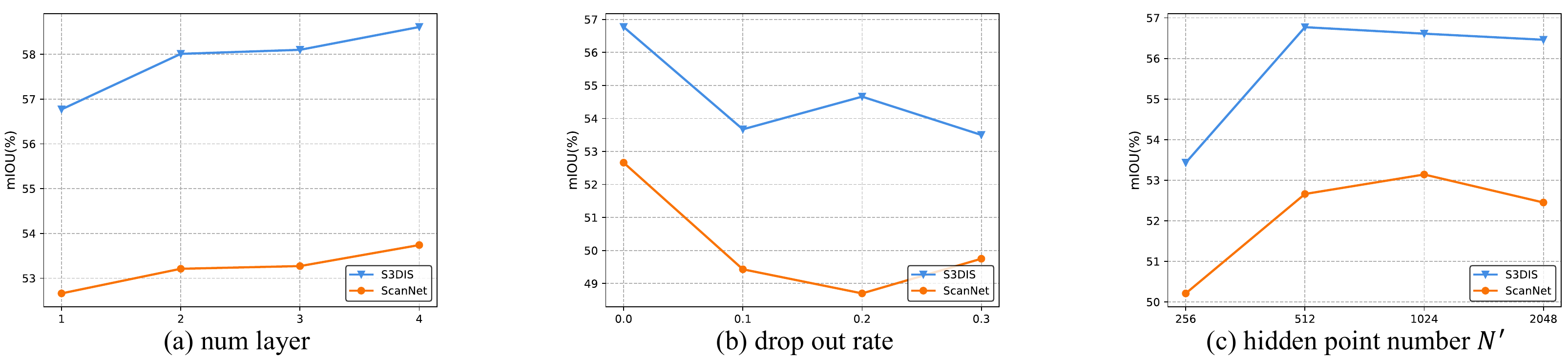}
	\caption{\small{Effects of three hyper-parameters, number of layers, drop out rate, and hidden point number $N'$, under \textbf{2-way 1-shot} setting on S3DIS (S$^0$) and ScanNet (S$^0$) datasets.}}
	\label{fig:hyper-parameters}
\end{figure*}

\subsection{Ablation Study}

\paragraph{Component Analysis}

We conduct detailed component ablation studies to evaluate each of our proposed modules in \tablename~\ref{ablation}. The experiments are conducted on both S3DIS and ScanNet under 2-way 1-shot and 2-way 5-shot settings. We adopt ProtoNet~\cite{garcia2017few, PANet} as our baseline. 
First of all, when adding the proposed Query-Guided Prototype Adaption (QGPA) module on our baseline, a performance gain of 3.97\% and 10.66\% in terms of mIoU under 1-shot settings is observed over the baseline on S3DIS and ScanNet, respectively, as shown in \tablename~\ref{ablation}. The performance gain is because of the benefit of our effective transformer design and the adaption of prototypes from support feature space to query feature space.
The superior result demonstrates that the capacity of transformer in adapting feature channel-wise correlations between samples, which is important in point cloud scenery, especially for few-shot learning with only a handful of training samples.

Then, by introducing Self-Reconstruction (SR) as auxiliary supervision, we further obtain significant improvement over the QGPA, \eg, 3.78\% and 4.66\% performance gain under 2-way 1-shot setting on S3DIS and ScanNet, respectively, as shown in the last row in \tablename~\ref{ablation}. The proposed Self-Reconstruction forces the prototypes to restore the support information generating them, which give constraints on the prototypes to retrain the class-related clues. With SR, better prototypes that contain discriminative feature representations are produced. Meantime, the gradients from QGPA may make the class-related support clues less pronounced while the proposed SR protect and enhance such clues. Meanwhile, we observe that the performance gain by Self-Reconstruction over baseline without QGPA is less than with QGPA, \eg, 0.17\% and 0.11\% under 2-way-1-shot setting on S3DIS and ScanNet, respectively, as shown in the third row in \tablename~\ref{ablation}.
This phenomenon suggests that original prototypes extract discriminative clues from support samples well and adding extra constraints on this basis does not play a big role. When combining these two modules together, our full method achieves the best results that improve substantially over the baseline, which demonstrates that the proposed SR and QGPA are mutually beneficial. 
It is worth noting that when SR is added on the model with QGPA, it can play a better positive role than on the model without QGPA, which is in line with our original motivation. 
When we utilize QGPA to align prototypes with the query feature distribution, it may lose the original discriminative and semantic characteristics deviate from the original prototype. With our proposed SR as supervision, prototypes can preserve these informative clues by reconstructing support gourd-truth masks from themselves.

\paragraph{Ablation Study for Baseline Configuration}
We study the effects of various designs of the ProtoNet~\cite{garcia2017few} since it is the baseline of our method. 
The results of different variants are listed in \tablename~\ref{tab:ablation-baseline}. The results are improved consistently with the help of data augmentation (AUG), multi-scale feature aggregation (MS)~\cite{FPN,CGBNet}, and align loss (AL)~\cite{PANet}. As shown in \tablename~\ref{tab:ablation-baseline}, based on the vanilla ProtoNet~\cite{garcia2017few}, with random scale and random shift to augment training samples, an improvement of 0.81\% on S3DIS and 2.2\% on ScanNet is achieved. Then, by aggregating multi-scale features from the backbone DGCNN~\cite{wang2019dynamic}, we further improve the results by 2.05\% on S3DIS and 3.33\% on ScanNet. Then by introducing align loss~\cite{PANet} with a performance gain of around 1\%, the baseline achieves 52.17\% on S3DIS and 41.03\% on ScanNet.

\begin{figure}[t]
	\centering
	\includegraphics[scale=0.5]{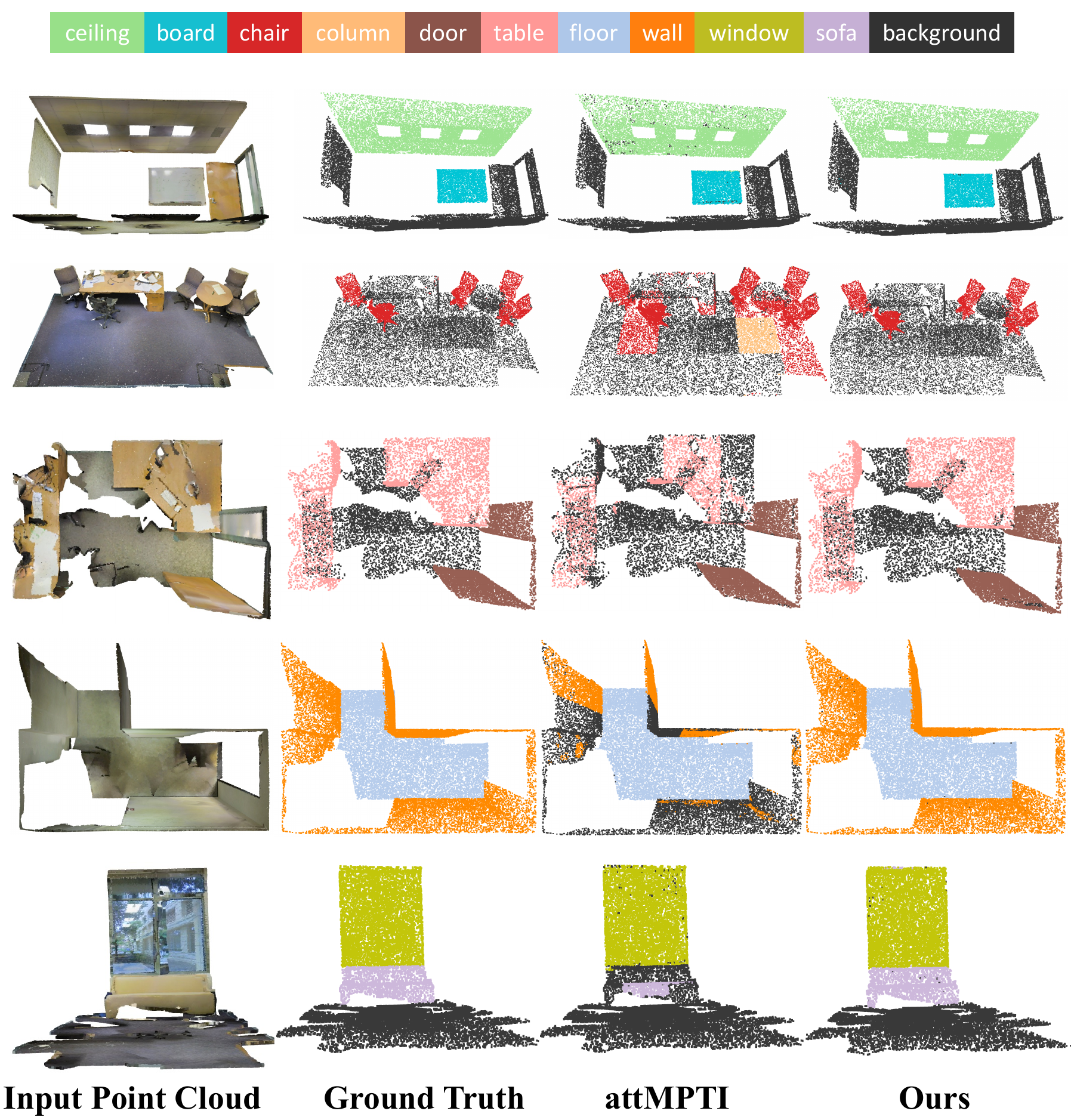} 
	\caption{Qualitative results of our method in \textbf{2-way 1-shot} point cloud few-shot segmentation on the \textbf{S3DIS} dataset in comparison to the ground truth and attMPTI~\cite{zhao2021few}. Five combinations of  \textit{2-way} are illustrated from the top to bottom rows, \textit{i.e.}, ``\textit{ceiling, board}'' (first row), ``\textit{chair, column}'' (second row), ``\textit{door, table}'' (third row), ``\textit{floor, wall}'' (fourth row).``\textit{window, sofa}'' (last row).}
	\label{fig:s3dis-visualization}
\end{figure}

\paragraph{Ablation Study for QGPA Configuration}
In \figurename~\ref{fig:hyper-parameters}, we illustrate the effects of three hyper-parameters in our proposed QGPA configuration (\ie, number of transformer block layer, dropout rate, hidden point number $N'$) under the setting of 2-way 1-shot point cloud semantic segmentation on one split of S3DIS and ScanNet. 
As shown in \figurename~\ref{fig:hyper-parameters}~\!(a), increasing the layers of QGPA achieves better results, but overly large layers consume more computing resources and slow down the inference speed. Thus, we choose a single layer to achieve a good balance of accuracy and efficiency.
As \figurename~\ref{fig:hyper-parameters}~\!(b) reveals that increasing drop out rate degrades the result a lot, the drop rate of 0.0 gives the best result on both datasets. 
While in \figurename~\ref{fig:hyper-parameters}~\!(c), with the increase of hidden point number, the result first rises and then becomes flat. Therefore, our transformer's hidden point number $N'$ is set to 512 to achieve robust performance and keep efficient.

\paragraph{Comparison with Other Designs for QGPA}
To verify the superiority of our proposed QGPA, we list several other SOTA transformer-like modules design in \tablename~\ref{tab:transformer-variations}.
For a fair comparison, except for the transformer design, these methods are trained with the same experimental configuration and we conduct experiments based on the same baseline without Self-Reconstruction.
Classifier Weight Transformer (CWT)~\cite{CWT} is a few-shot segmentation transformer architecture which modifies classifier weight adapting to each query feature. Classifier weight can be regarded as prototypes, hence here we set the prototype as \textit{query}; query feature as the \textit{key} and \textit{value} input of CWT transformer, please refer to CWT's paper for CWT architecture details. 
Besides, we use DETR~\cite{detr} decoder branch which contains self-and cross-attention block design. We set the prototype as \textit{query}; query feature as the \textit{key} and \textit{value}, similar with CWT.
It is worth noting that the number of prototypes varies with the number of ways $C$. For instance, in a 2-way setting, there would be three prototypes. Therefore, the self-attention block in DETR is able to work as intended.
As the experimental results shown in \tablename~\ref{tab:transformer-variations}, CWT and the vanilla DETR transformer perform inferior to our proposed Query-Guided Prototype Adaption (QGPA), where performance drops of -3.98\%/-6.5\% on S3DIS and -9.80\%/-7.66\% on ScanNet under 1-shot settings are observed, respectively. This confirms that simply applying Transformers for the 3D point cloud segmentation task is not effective, because they ignore the discrepancy in the distribution of features in the channel-wise space.
D-QGPA represents the degradation of our proposed QGPA which replaces support feature with query feature for \textit{key} input. It will lead to that attention value is calculated from query feature itself and cannot get knowledge from support feature. 
We find that D-QGPA does not lead to improvement but a large drop (-14.18\% under 1-shot On ScanNet) compared to QGPA. This indicates the essential importance of adaption from support to query feature space. Without support feature, the prototype adaption lacks the information of the source, only the information of the target, which makes the adaptation process out of action.
The model parameters and inference speed of these methods are also listed in \tablename~\ref{tab:transformer-variations}. As can be seen, our method is more lightweight and efficient.

\begin{figure}[t]
	\centering
	\includegraphics[scale=0.5]{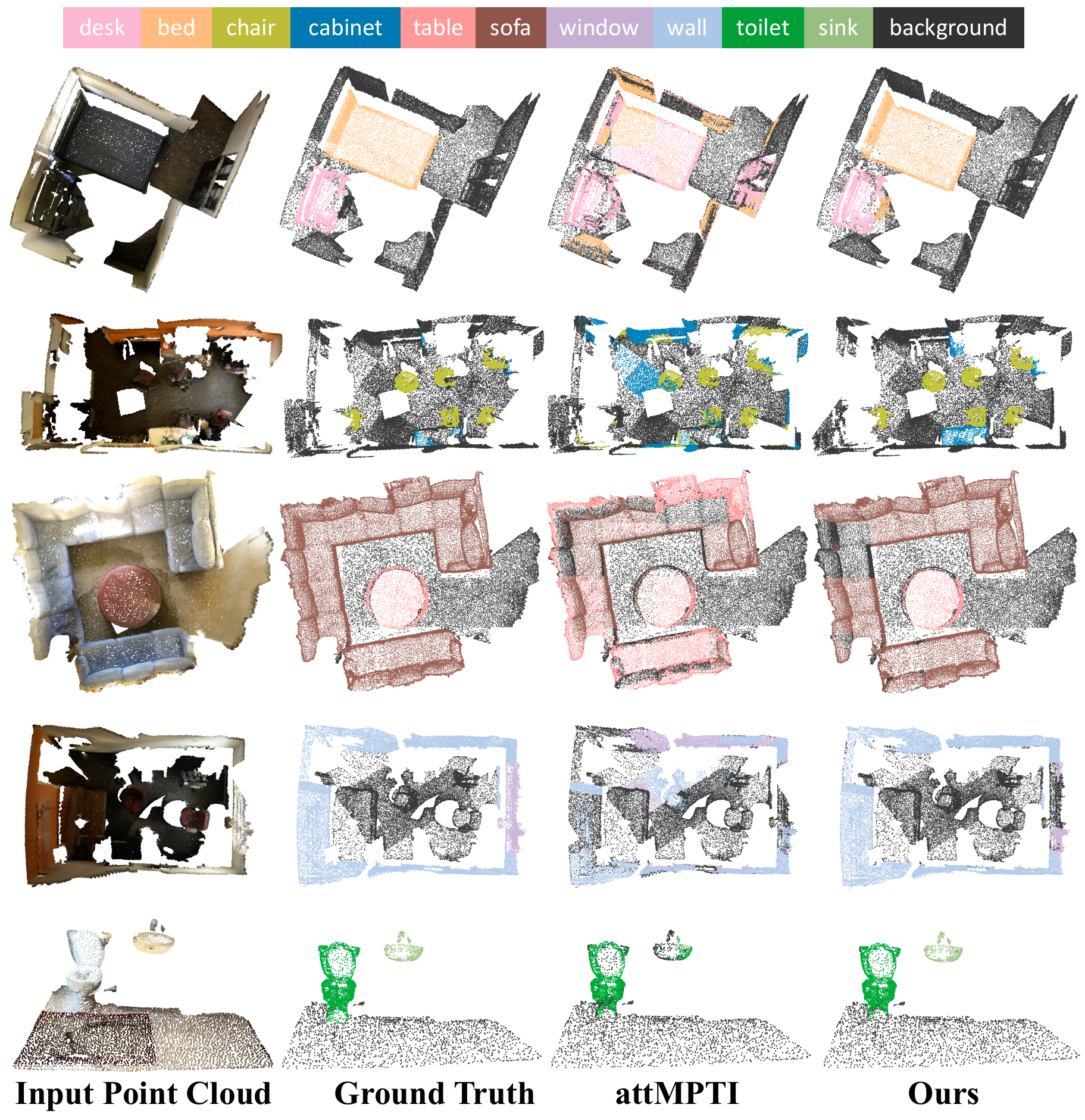} 
	\caption{Qualitative results of our method in \textbf{2-way~1-shot} point cloud few-shot semantic segmentation setting on the \textbf{ScanNet} dataset in comparison to the ground truth and attMPTI~\cite{zhao2021few}. Five combinations of  \textit{2-way} are illustrated from the top to bottom rows, \ie, ``\textit{desk, bed}'' (first row), ``\textit{chair, cabinet}'' (second row), ``\textit{table, sofa}'' (third row), ``\textit{window, wall}'' (fourth row), ``\textit{toilet, sink}'' (last row).}
	\label{fig:ScanNet-visualization}
\end{figure}

\begin{table}[t]
\footnotesize
	\begin{center}
	\setlength\tabcolsep{8pt}
	\renewcommand\arraystretch{1.3}
		\begin{tabular}{ C{1.0cm} | C{1.0cm} C{1.0cm} | C{1.0cm} C{1.0cm} }
			\specialrule{.1em}{.05em}{.05em}
			\multirow{2}{*}{Type}
			& \multicolumn{2}{c?}{\textbf{S3DIS}}
			& \multicolumn{2}{c}{\textbf{ScanNet}} \\ 
			\cline{2-5} 
			& 1-shot &5-shot&1-shot & 5-shot
			\\ \hline
		Original &62.76  &67.85   &56.51 &62.10     \\
		Refined & 59.85& 66.03  & 52.46  & 59.97\\
			\specialrule{.1em}{.05em}{.05em}
		\end{tabular}
	\caption{\small{Effect of different choices (original or refined prototype) to conduct Self-Reconstruction under \textbf{2-way}
setting on \textbf{S3DIS} and \textbf{ScanNet} dataset using mean-IoU metric (\%). }\label{ablation_SR}}
\end{center}
\end{table}

\begin{table*}[htbp]
\small
	\centering
	\renewcommand\arraystretch{1.2}
	\setlength\tabcolsep{9pt}
	\scalebox{0.996}{
		\begin{tabular}{L{2.2cm}?C{0.6cm} C{0.6cm} C{0.6cm}? C{0.6cm} C{0.6cm} C{0.6cm} ? C{0.6cm}  C{0.6cm}  C{0.6cm} ? C{0.6cm} C{0.6cm} C{0.6cm}}
			\specialrule{.1em}{.05em}{.05em}
			\multirow{3}{*}{\textbf{Method}} 
			& \multicolumn{6}{c?}{\textbf{2-way}}
			& \multicolumn{6}{c}{\textbf{3-way}} \\ \cline{2-13} 
			& \multicolumn{3}{c?}{\textbf{1-shot}} & \multicolumn{3}{c?}{\textbf{5-shot}}
			& \multicolumn{3}{c?}{\textbf{1-shot}} & \multicolumn{3}{c}{\textbf{5-shot}}
			\\ \cline{2-13} 
			& \multicolumn{1}{c}{S$^0$} & \multicolumn{1}{c}{S$^1$} &\multicolumn{1}{c?}{mean} 
			& \multicolumn{1}{c}{S$^0$} & \multicolumn{1}{c}{S$^1$} 
			& \multicolumn{1}{c?}{mean}              
			& \multicolumn{1}{c}{S$^0$} & \multicolumn{1}{c}{S$^1$} &\multicolumn{1}{c?}{mean} 
			& \multicolumn{1}{c}{S$^0$} & \multicolumn{1}{c}{S$^1$} 
			& \multicolumn{1}{c}{mean}  
			\\  \hline
			FT~\cite{zhao2021few}  & 36.34 & 38.79  & 37.57  & 56.49  & 56.99  & 56.74  & 30.05  & 32.19 & 31.12  & 46.88  & 47.57  &  47.23 \\ 
			ProtoNet~\cite{zhao2021few}  & 48.39 & 49.98  &  49.19 &  57.34 & 63.22  & 60.28  & 40.81 & 45.07  &  42.94 & 49.05  & 53.42  & 51.24 \\ 
			AttProtoNet~\cite{zhao2021few}  & 50.98 & 51.90  &  51.44 & 61.02  & 65.25  & 63.14  & 42.16 & 46.76  &  44.46 & 52.20  & 56.20  & 54.20\\
			MPTI~\cite{zhao2021few}  & 52.27 &  51.48 &  51.88 & 58.93  & 60.56  & 59.75  & 44.27 & 46.92  &  45.60 & 51.74  & 48.57  & 50.16\\
			attMPTI~\cite{zhao2021few}  & 53.77 & 55.94  & 54.86  & 61.67  & 67.02  & 64.35 & 45.18 & 49.27  & 47.23  &  54.92 &  56.79 & 55.86 \\
			\hline
			\textbf{Ours}  & \textbf{59.45} &  \textbf{66.08}& \textbf{62.76}  &  \textbf{65.40} & \textbf{70.30}  & \textbf{67.85}  & \textbf{48.99} &  \textbf{56.57}  &\textbf{52.78}   & \textbf{61.27}  &\textbf{60.81}   &\textbf{61.04} \\

			\specialrule{.1em}{.05em}{.05em}
		\end{tabular}
	}
	\caption{\small{Results on \textbf{S3DIS} \cite{armeni20163d} dataset using mean-IoU metric (\%). S$^i$ denotes the split $i$ is used for testing.}}

	\label{tab:exp-s3dis}
\end{table*}

\begin{table*}[htbp]\small
\renewcommand\arraystretch{1.2}
\setlength\tabcolsep{10pt}
	\centering
	\scalebox{0.96}{
		\begin{tabular}{L{2.2cm}?C{0.6cm} C{0.6cm} C{0.6cm}? C{0.6cm} C{0.6cm} C{0.6cm} ? C{0.6cm}  C{0.6cm}  C{0.6cm} ? C{0.6cm} C{0.6cm} C{0.6cm}}
			\specialrule{.1em}{.05em}{.05em}
			\multirow{3}{*}{\textbf{Method}} 
			& \multicolumn{6}{c?}{\textbf{2-way}}
			& \multicolumn{6}{c}{\textbf{3-way}} \\ \cline{2-13} 
			& \multicolumn{3}{c?}{\textbf{1-shot}} & \multicolumn{3}{c?}{\textbf{5-shot}}
			& \multicolumn{3}{c?}{\textbf{1-shot}} & \multicolumn{3}{c}{\textbf{5-shot}}
			\\ \cline{2-13} 
			& \multicolumn{1}{c}{S$^0$} & \multicolumn{1}{c}{S$^1$} &\multicolumn{1}{c?}{mean} 
			& \multicolumn{1}{c}{S$^0$} & \multicolumn{1}{c}{S$^1$} 
			& \multicolumn{1}{c?}{mean}              
			& \multicolumn{1}{c}{S$^0$} & \multicolumn{1}{c}{S$^1$} &\multicolumn{1}{c?}{mean} 
			& \multicolumn{1}{c}{S$^0$} & \multicolumn{1}{c}{S$^1$} 
			& \multicolumn{1}{c}{mean}  
			\\  \hline
			FT~\cite{zhao2021few}  & 31.55 & 28.94  & 30.25  &  42.71 & 37.24  & 39.98  & 23.99 & 19.10  & 21.55  & 34.93  &  28.10 & 31.52 \\ 
			ProtoNet~\cite{zhao2021few}  & 33.92 &  30.95 & 32.44  & 45.34  & 42.01  & 43.68  & 28.47 &  26.13 &  27.30 &   37.36 &  34.98 & 36.17 \\ 
			AttProtoNet~\cite{zhao2021few}  & 37.99 & 34.67  &  36.33 & 52.18  &  46.89 & 49.54  & 32.08 & 28.96  & 30.52 & 44.49  & 39.45  & 41.97 \\ 
			MPTI~\cite{zhao2021few}  & 39.27 & 36.14  & 37.71  &  46.90 & 43.59  & 45.25  & 29.96 & 27.26  & 28.61  & 38.14  & 34.36  & 36.25\\  
			attMPTI~\cite{zhao2021few}  & 42.55 & 40.83  &  41.69 & 54.00  & 50.32  & 52.16  & 35.23 & 30.72  & 32.98 & 46.74 & 40.80  & 43.77\\ 
			\hline
			\textbf{Ours}  & \textbf{57.08} & \textbf{55.94}  & \textbf{56.51}  &  \textbf{64.55} & \textbf{59.64}  & \textbf{62.10}  & \textbf{55.27} &  \textbf{55.60} & \textbf{55.44} & \textbf{59.02}  &  \textbf{53.16} &\textbf{56.09} \\

			\specialrule{.1em}{.05em}{.05em}
		\end{tabular}
	}
	\caption{\small{Results on \textbf{ScanNet}~\cite{dai2017scannet} dataset using mean-IoU metric (\%). S$^i$ denotes the split $i$ is used for testing.}}
	\label{tab:exp-scannet}
\end{table*}

\paragraph{Ablation Study for Different Choices of Prototypes in Self-Reconstruction}
Upon transferring the prototype to the query feature space, the feature gap between the refined prototype and the original support feature arises.
Consequently, utilizing the refined prototype in query feature space to reconstruct the support mask in support feature space is unreasonable. In contrast, using constraints on the original prototype can facilitate the extraction of a more discriminative representation from the support set. This serves as a foundation for the subsequent adaptation's success, allowing the refined prototype to indirectly retain critical class and semantic information. Furthermore, we conduct extensive experiments to assess the choice of original prototypes or refined prototypes in conducting reconstructed support masks, as shown in \tablename~\ref{ablation_SR}.
Utilizing the original prototypes yields superior performance, exhibiting an increase of up to 4.05\% in mIoU compared to the refined prototype, which is in accordance with our previous analysis.

\subsection{Qualitative Result}
To qualitatively demonstrate the performance of our proposed approach, we visualize some point cloud segmentation result on the S3DIS and ScanNet in \figurename~\ref{fig:s3dis-visualization} and \figurename~\ref{fig:ScanNet-visualization}, respectively. 
In both \figurename~\ref{fig:s3dis-visualization} and \figurename~\ref{fig:ScanNet-visualization}, the first column is visualization of input point clouds, the second column is ground-truth masks, the third and fourth columns are predicted masks by attMPTI~\cite{zhao2021few} and our proposed approach, respectively. Our approach achieves better results than attMPTI~\cite{zhao2021few}. For example, in the last row of \figurename~\ref{fig:s3dis-visualization}, our approach segments the ``\textit{sofa}'' very well, while attMPTI identifies part of the ``\textit{sofa}'' to ``\textit{background}''. In the last row of \figurename~\ref{fig:ScanNet-visualization}, attMPTI incorrectly classifies ``\textit{sink}'' to ``\textit{toilet}'' and ``\textit{background}'', while our approach generates high-quality mask for ``\textit{sink}''. The qualitative results in \figurename~\ref{fig:s3dis-visualization} and \figurename~\ref{fig:ScanNet-visualization} demonstrate the effectiveness of our proposed approach and our approach's superior over previous state-of-the-art method attMPTI~\cite{zhao2021few}.

\subsection{Comparison with State-of-the-Art Methods}

In \tablename~\ref{tab:exp-s3dis} and \tablename~\ref{tab:exp-scannet}, we compare with previous state-of-the-art methods and report our quantitative results on S3DIS~\cite{armeni20163d} and ScanNet~\cite{dai2017scannet} datasets, respectively. Our proposed method significantly outperforms previous state-of-the-art method by a large margin. We outperforms attMPTI~\cite{zhao2021few} in all settings. For example, our proposed approach is 7.90\% and 3.5\% better than attMPTI~\cite{zhao2021few} under 2-way 1-shot and 2-way 5-shot settings on S3DIS, and is 22.46\% and 12.32\% better than attMPTI under 3-way 1shot, 3-way 5-shot settings on ScanNet.
Compared to ProtoNet~\cite{garcia2017few} which has a similar design paradigm to us, our method achieves up to 13.57\% and 24.07\% gains on S3DIS and ScanNet, respectively. The huge improvements demonstrate that our method can obtain more dicriminative and adaptive prototypes from not only support samples but also support-query feature adaption.
The superior results obtained by our method show that the intra-class-sample-variations problem is critical in 3D point cloud scenery, and our proposed Query-Guided Prototype Adaption and Self-Reconstruction are effective to address this problem.

\subsection{Comparison with State-of-the-Art Zero-shot Methods}
We further evaluate our model with semantic prototype projection branch, as shown in \tablename~\ref{tab:exp-3d}. During testing, the point-level annotations of the support set are replaced by semantic prototypes which are generated from our semantic branch. We report our results with both word2vec~\cite{word2vector} and CLIP~\cite{clip, zhang2021pointclip} as text encoder. 
On the one hand,
Compared with \tablename~\ref{tab:exp-s3dis}, our text-based model achieves competitive results compared to the ones with visual support samples. 
Therefore, the introduction of semantic projection network is able to bridge the gap between visual support and semantic words and establish a generalized framework for few- and zero-shot learning that achieves superior performance regardless of whether the input is in the form of semantic words or visual support samples.
It is worth noting that our zero-shot segmentation model achieves better results under 5-shot training than 1-shot. This is because we jointly train our few-shot model and zero-shot model, and use the visual prototypes from support samples as the ground truth of our word-projected prototypes during training. More support samples produce better visual prototypes, which contribute to training a better text-vision projection network that can generate more accurate word-projected prototypes.
On the other hand, to provide a fair comparison with other zero-shot methods, we follow their official code and evaluate their method under our experimental settings for data augmentations for the training and the training and testing subset splitting of the data to get the results in \tablename~\ref{tab:exp-3d}. The results show that our method has significant improvement over 3DGenZ~\cite{3DGenZ}, the only open-source 3D zero-shot segmentation method to the best of our knowledge. This comparison further validates the effectiveness of our method.

\begin{table*}[htbp]\small
\renewcommand\arraystretch{1.2}
\setlength\tabcolsep{10pt}
	\centering
	\scalebox{0.96}{
		\begin{tabu}{L{1.0cm} L{1.5cm}?C{0.4cm} C{0.4cm} ? C{0.4cm} C{0.4cm}  ? C{0.4cm}  C{0.4cm}   ? C{0.4cm} C{0.4cm}}
			\specialrule{.1em}{.05em}{.05em}
			&
			& \multicolumn{4}{c?}{\textbf{S3DIS}}
			& \multicolumn{4}{c}{\textbf{ScanNet}} 
                \\ \cline{3-10} 
			&& \multicolumn{2}{c?}{\textbf{2-way}} & \multicolumn{2}{c?}{\textbf{3-way}}
			& \multicolumn{2}{c?}{\textbf{2-way}} & \multicolumn{2}{c}{\textbf{3-way}}
			\\ \cline{3-10} 
			\multirow{-3}{*}{Embed}&\multirow{-3}{*}{Method}& \multicolumn{1}{c}{1-shot} & \multicolumn{1}{c|}{5-shot} 
			& \multicolumn{1}{c}{1-shot} & \multicolumn{1}{c|}{5-shot}          
			& \multicolumn{1}{c}{1-shot} & \multicolumn{1}{c|}{5-shot} 
			& \multicolumn{1}{c}{1-shot} & \multicolumn{1}{c}{5-shot} 
			\\  \hline
   word2vec&3DGenZ~\cite{3DGenZ} & 34.93 & 36.12 &23.08 & 27.52& 29.07&31.65 & 28.13& 28.01\\

word2vec&	\textbf{Ours}  & \textbf{59.98} & \textbf{63.54}  & \textbf{48.91}  &  \textbf{55.62} & \textbf{54.72}  & \textbf{58.94}  & \textbf{53.78} &  \textbf{53.50} \\
CLIP&	\textbf{Ours} & \textbf{61.09} & \textbf{64.91}  & \textbf{50.18}  &  \textbf{59.10} & \textbf{56.12}  & \textbf{60.65}  & \textbf{55.24} &  \textbf{55.04}  \\
			\specialrule{.1em}{.05em}{.05em}
		\end{tabu}
	}
	\caption{\small{Comparison of zero-shot 3D semantic segmentation results on \textbf{S3DIS} and \textbf{ScanNet} dataset using mean-IoU (\%).}}
	\label{tab:exp-3d}
\end{table*}

\subsection{Computational Complexity}
In \tablename~\ref{tab:r4q4}, we present the number of parameters and computational complexity of our proposed model and previous SOTA method attMPTI~\cite{zhao2021few}. The Query-Guided Prototype Adaption (QGPA) introduces two linear layers that map the point cloud from 2048 to 512, resulting in a moderate increase in the number of parameters.
With the addition of Self-Reconstruction (SR), we only need to calculate an additional loss item, and no additional parameters are introduced, thus keeping the computational complexity unchanged.
Finally, we integrate our Semantic Project Network (SPN) to obtain the final model. As we only need to learn the mapping from semantic words to visual prototypes, the increase in the number of parameters and computational complexity is minimal.
Our model demonstrates strong performance while maintaining a relatively low number of parameters and computational complexity, particularly in terms of FPS. Although attMPTI's Transductive Inference process doesn't increase the parameter count, it significantly slows down inference speed. As a result, our approach is a highly effective and efficient solution for few- and zero-shot 3D point cloud semantic segmentation, delivering superior results and faster FPS.

\begin{table}[h] 
\footnotesize
	\centering
	\renewcommand\arraystretch{1.2}
		\begin{tabular}{C{0.5cm} C{0.5cm} C{0.5cm}| C{0.8cm}  C{0.8cm}C{0.8cm} |C{0.8cm}C{0.8cm}}
			\specialrule{.1em}{.05em}{.05em}

			QGPA &SR &SPN &FLOPs & \#Params. &FPS &S3DIS &ScanNet \\ \hline 
      		      \multicolumn{3}{c|}{attMPTI~\cite{zhao2021few}}&  7.72G & 357.82K & 1.6& 54.86&41.69\\
                  \hline 
   		    \xmarkg&\xmarkg &\xmarkg  &  7.12G & 352.19K & 33.3 & 55.01 & 41.19\\
		    \cmark&\xmarkg &\xmarkg  &  7.48G & 2.48M  & 33.3 & 58.98 &51.85\\
       	    \cmark&  \cmark&\xmarkg &  7.48G  & 2.48M &33.3  &62.76 &56.51\\
		      \cmark& \cmark & \cmark& 7.49G  & 2.79M & 33.3 & 61.09 &56.12\\

			\specialrule{.1em}{.05em}{.05em}
   
	\end{tabular}
	\caption{\small{Analysis of computation cost of the proposed method and the results are under the 2-way 1-shot setting.}}
	\label{tab:r4q4}
\end{table}

\section{Conclusion}
We propose a prototype adaption and projection network for few-shot and zero-shot point cloud semantic segmentation. By analyzing the feature channel distribution of 2D images and 3D point clouds, we have observed that the feature intra-class variation of 3D point clouds is worse than 2D due to the lack of pre-training on large-scale datasets. We hence propose a Query-Guided Prototype Adaption (QGPA) module to map the prototypes extracted in support feature space to the query feature space, which greatly improves the few-shot segmentation performance. To preserve more class-specific clues in prototypes, we introduce Self-Reconstruction (SR) that enables the prototype to reconstruct the corresponding mask as well as possible. Furthermore, a semantic projection network is proposed to deal with the zero-shot learning cases where no annotated sample is provided but just category names. The semantic projection network makes our model more practical in the real-world. We evaluate the proposed approach on two popular 3D point cloud segmentation datasets, which show new state-of-the-art performances with significant improvement over previous methods.

\ifCLASSOPTIONcaptionsoff
  \newpage
\fi



{\small
\bibliographystyle{IEEEtran}
\bibliography{reference}

\begin{thebibliography}{10}
\providecommand{\url}[1]{#1}
\csname url@samestyle\endcsname
\providecommand{\newblock}{\relax}
\providecommand{\bibinfo}[2]{#2}
\providecommand{\BIBentrySTDinterwordspacing}{\spaceskip=0pt\relax}
\providecommand{\BIBentryALTinterwordstretchfactor}{4}
\providecommand{\BIBentryALTinterwordspacing}{\spaceskip=\fontdimen2\font plus
\BIBentryALTinterwordstretchfactor\fontdimen3\font minus
  \fontdimen4\font\relax}
\providecommand{\BIBforeignlanguage}[2]{{%
\expandafter\ifx\csname l@#1\endcsname\relax
\typeout{** WARNING: IEEEtran.bst: No hyphenation pattern has been}%
\typeout{** loaded for the language `#1'. Using the pattern for}%
\typeout{** the default language instead.}%
\else
\language=\csname l@#1\endcsname
\fi
#2}}
\providecommand{\BIBdecl}{\relax}
\BIBdecl

\bibitem{landrieu2017large}
L.~Landrieu and M.~Simonovsky, ``Large-scale point cloud semantic segmentation
  with superpoint graphs,'' in \emph{Proc. IEEE Conf. Comput. Vis. Pattern
  Recognit.}, 2018, pp. 4558--4567.

\bibitem{qi2017pointnet}
C.~R. Qi, H.~Su, K.~Mo, and L.~J. Guibas, ``Pointnet: Deep learning on point
  sets for 3d classification and segmentation,'' in \emph{Proc. IEEE Conf.
  Comput. Vis. Pattern Recognit.}, 2017, pp. 652--660.

\bibitem{huang2018recurrent}
Q.~Huang, W.~Wang, and U.~Neumann, ``Recurrent slice networks for 3d
  segmentation of point clouds,'' in \emph{Proc. IEEE Conf. Comput. Vis.
  Pattern Recognit.}, 2018, pp. 2626--2635.

\bibitem{li2018pointcnn}
Y.~Li, R.~Bu, M.~Sun, W.~Wu, X.~Di, and B.~Chen, ``Pointcnn: Convolution on
  x-transformed points,'' in \emph{Proc. Adv. Neural Inform. Process. Syst.},
  2018, pp. 820--830.

\bibitem{li2022primitive3d}
X.~Li, H.~Ding, Z.~Tong, Y.~Wu, and Y.~M. Chee, ``Primitive3d: 3d object
  dataset synthesis from randomly assembled primitives,'' in \emph{Proc. IEEE
  Conf. Comput. Vis. Pattern Recognit.}, 2022, pp. 15\,947--15\,957.

\bibitem{wang2019dynamic}
Y.~Wang, Y.~Sun, Z.~Liu, S.~E. Sarma, M.~M. Bronstein, and J.~M. Solomon,
  ``Dynamic graph cnn for learning on point clouds,'' \emph{ACM Transactions on
  Graphics (TOG)}, vol.~38, no.~5, pp. 1--12, 2019.

\bibitem{li2023transformer}
X.~Li, H.~Ding, W.~Zhang, H.~Yuan, J.~Pang, G.~Cheng, K.~Chen, Z.~Liu, and
  C.~C. Loy, ``Transformer-based visual segmentation: A survey,'' \emph{arXiv
  preprint arXiv:2304.09854}, 2023.

\bibitem{zhao2021few}
N.~Zhao, T.-S. Chua, and G.~H. Lee, ``Few-shot 3d point cloud semantic
  segmentation,'' in \emph{Proc. IEEE Conf. Comput. Vis. Pattern Recognit.},
  2021, pp. 8873--8882.

\bibitem{PANet}
K.~Wang, J.~H. Liew, Y.~Zou, D.~Zhou, and J.~Feng, ``Panet: Few-shot image
  semantic segmentation with prototype alignment,'' in \emph{International
  Conference on Computer Vision}.\hskip 1em plus 0.5em minus 0.4em\relax
  {IEEE}, 2019, pp. 9196--9205.

\bibitem{liu2020part}
Y.~Liu, X.~Zhang, S.~Zhang, and X.~He, ``Part-aware prototype network for
  few-shot semantic segmentation,'' in \emph{Proc. Eur. Conf. Comput.
  Vis.}\hskip 1em plus 0.5em minus 0.4em\relax Cham: Springer International
  Publishing, 2020, pp. 142--158.

\bibitem{liu2022few}
W.~Liu, C.~Zhang, H.~Ding, T.-Y. Hung, and G.~Lin, ``Few-shot segmentation with
  optimal transport matching and message flow,'' \emph{IEEE Trans. Multimedia},
  2022.

\bibitem{yang2020prototype}
B.~Yang, C.~Liu, B.~Li, J.~Jiao, and Q.~Ye, ``Prototype mixture models for
  few-shot semantic segmentation,'' in \emph{Proc. Eur. Conf. Comput.
  Vis.}\hskip 1em plus 0.5em minus 0.4em\relax Cham: Springer International
  Publishing, 2020, pp. 763--778.

\bibitem{deng2009imagenet}
J.~Deng, W.~Dong, R.~Socher, L.-J. Li, K.~Li, and L.~Fei-Fei, ``Imagenet: A
  large-scale hierarchical image database,'' in \emph{Proc. IEEE Conf. Comput.
  Vis. Pattern Recognit.}, 2009, pp. 248--255.

\bibitem{graham20183d}
B.~Graham, M.~Engelcke, and L.~Van Der~Maaten, ``3d semantic segmentation with
  submanifold sparse convolutional networks,'' in \emph{Proc. IEEE Conf.
  Comput. Vis. Pattern Recognit.}, 2018, pp. 9224--9232.

\bibitem{SubmanifoldSparseConvNet}
B.~Graham and L.~van~der Maaten, ``Submanifold sparse convolutional networks,''
  \emph{arXiv preprint arXiv:1706.01307}, 2017.

\bibitem{choy20194d}
C.~Choy, J.~Gwak, and S.~Savarese, ``4d spatio-temporal convnets: Minkowski
  convolutional neural networks,'' in \emph{Proc. IEEE Conf. Comput. Vis.
  Pattern Recognit.}, 2019, pp. 3075--3084.

\bibitem{qi2017pointnet++}
C.~R. Qi, L.~Yi, H.~Su, and L.~J. Guibas, ``Pointnet++: Deep hierarchical
  feature learning on point sets in a metric space,'' in \emph{Proc. Adv.
  Neural Inform. Process. Syst.}, 2017, pp. 5099--5108.

\bibitem{wu2019pointconv}
W.~Wu, Z.~Qi, and L.~Fuxin, ``Pointconv: Deep convolutional networks on 3d
  point clouds,'' in \emph{CVPR}, 2019, pp. 9621--9630.

\bibitem{zhao2020point}
H.~Zhao, L.~Jiang, J.~Jia, P.~H. Torr, and V.~Koltun, ``Point transformer,'' in
  \emph{ICCV}, 2021, pp. 16\,259--16\,268.

\bibitem{lai2022stratified}
X.~Lai, J.~Liu, L.~Jiang, L.~Wang, H.~Zhao, S.~Liu, X.~Qi, and J.~Jia,
  ``Stratified transformer for 3d point cloud segmentation,'' in \emph{CVPR},
  2022, pp. 8500--8509.

\bibitem{vu2022softgroup}
T.~Vu, K.~Kim, T.~M. Luu, X.~T. Nguyen, and C.~D. Yoo, ``Softgroup for 3d
  instance segmentation on point clouds,'' in \emph{CVPR}, 2022, pp.
  2708--2717.

\bibitem{snell2017prototypical}
J.~Snell, K.~Swersky, and R.~Zemel, ``Prototypical networks for few-shot
  learning,'' in \emph{Proc. Adv. Neural Inform. Process. Syst.}, 2017, pp.
  4077--4087.

\bibitem{vinyals2016matching}
O.~Vinyals, C.~Blundell, T.~Lillicrap, D.~Wierstra \emph{et~al.}, ``Matching
  networks for one shot learning,'' in \emph{Proc. Adv. Neural Inform. Process.
  Syst.}, 2016, pp. 3630--3638.

\bibitem{sung2018learning}
F.~Sung, Y.~Yang, L.~Zhang, T.~Xiang, P.~H. Torr, and T.~M. Hospedales,
  ``Learning to compare: Relation network for few-shot learning,'' in
  \emph{Proc. IEEE Conf. Comput. Vis. Pattern Recognit.}, 2018, pp. 1199--1208.

\bibitem{zhang2020deepemd}
C.~Zhang, Y.~Cai, G.~Lin, and C.~Shen, ``Deepemd: Few-shot image classification
  with differentiable earth mover's distance and structured classifiers,'' in
  \emph{Proc. IEEE Conf. Comput. Vis. Pattern Recognit.}, 2020, pp.
  12\,203--12\,213.

\bibitem{finn2017model}
C.~Finn, P.~Abbeel, and S.~Levine, ``Model-agnostic meta-learning for fast
  adaptation of deep networks,'' in \emph{Proceedings of the 34th International
  Conference on Machine Learning-Volume 70}.\hskip 1em plus 0.5em minus
  0.4em\relax JMLR. org, 2017, pp. 1126--1135.

\bibitem{rusu2018meta}
A.~A. Rusu, D.~Rao, J.~Sygnowski, O.~Vinyals, R.~Pascanu, S.~Osindero, and
  R.~Hadsell, ``Meta-learning with latent embedding optimization,'' \emph{arXiv
  preprint arXiv:1807.05960}, 2018.

\bibitem{cai2018memory}
Q.~Cai, Y.~Pan, T.~Yao, C.~Yan, and T.~Mei, ``Memory matching networks for
  one-shot image recognition,'' in \emph{Proc. IEEE Conf. Comput. Vis. Pattern
  Recognit.}, 2018, pp. 4080--4088.

\bibitem{lampert2009learning}
C.~H. Lampert, H.~Nickisch, and S.~Harmeling, ``Learning to detect unseen
  object classes by between-class attribute transfer,'' in \emph{Proc. IEEE
  Conf. Comput. Vis. Pattern Recognit.}\hskip 1em plus 0.5em minus 0.4em\relax
  IEEE, 2009, pp. 951--958.

\bibitem{D2Zero}
S.~He, H.~Ding, and W.~Jiang, ``Semantic-promoted debiasing and background
  disambiguation for zero-shot instance segmentation,'' in \emph{Proc. IEEE
  Conf. Comput. Vis. Pattern Recognit.}, 2023, pp. 11\,238--11\,247.

\bibitem{zhang2021prototypical}
H.~Zhang and H.~Ding, ``Prototypical matching and open set rejection for
  zero-shot semantic segmentation,'' in \emph{Proc. IEEE Int. Conf. Comput.
  Vis.}, 2021, pp. 6974--6983.

\bibitem{PADing}
S.~He, H.~Ding, and W.~Jiang, ``Primitive generation and semantic-related
  alignment for universal zero-shot segmentation,'' in \emph{Proc. IEEE Conf.
  Comput. Vis. Pattern Recognit.}, 2023, pp. 11\,238--11\,247.

\bibitem{demirel2017attributes2classname}
B.~Demirel, R.~Gokberk~Cinbis, and N.~Ikizler-Cinbis, ``Attributes2classname: A
  discriminative model for attribute-based unsupervised zero-shot learning,''
  in \emph{Proceedings of the IEEE International Conference on Computer Vision
  (ICCV)}, 2017, pp. 1232--1241.

\bibitem{li2018deep}
Y.~Li, Z.~Jia, J.~Zhang, K.~Huang, and T.~Tan, ``Deep semantic structural
  constraints for zero-shot learning,'' in \emph{Proceedings of the AAAI
  Conference on Artificial Intelligence}, 2018, pp. 7049--7056.

\bibitem{gan2015exploring}
C.~Gan, M.~Lin, Y.~Yang, Y.~Zhuang, and A.~G. Hauptmann, ``Exploring semantic
  inter-class relationships (sir) for zero-shot action recognition,'' in
  \emph{Proceedings of the AAAI Conference on Artificial Intelligence}, 2015,
  pp. 3769--3775.

\bibitem{zhang2016zero}
Z.~Zhang and V.~Saligrama, ``Zero-shot learning via joint latent similarity
  embedding,'' in \emph{Proc. IEEE Conf. Comput. Vis. Pattern Recognit.}, 2016,
  pp. 6034--6042.

\bibitem{yu2013designing}
F.~X. Yu, L.~Cao, R.~S. Feris, J.~R. Smith, and S.-F. Chang, ``Designing
  category-level attributes for discriminative visual recognition,'' in
  \emph{Proc. IEEE Conf. Comput. Vis. Pattern Recognit.}, 2013, pp. 771--778.

\bibitem{ding2018context}
H.~Ding, X.~Jiang, B.~Shuai, A.~Q. Liu, and G.~Wang, ``Context contrasted
  feature and gated multi-scale aggregation for scene segmentation,'' in
  \emph{Proc. IEEE Conf. Comput. Vis. Pattern Recognit.}, 2018, pp. 2393--2402.

\bibitem{shuai2018toward}
B.~Shuai, H.~Ding, T.~Liu, G.~Wang, and X.~Jiang, ``Toward achieving robust
  low-level and high-level scene parsing,'' \emph{IEEE Transactions on Image
  Processing}, vol.~28, no.~3, pp. 1378--1390, 2018.

\bibitem{SVCNet}
H.~Ding, X.~Jiang, B.~Shuai, A.~Q. Liu, and G.~Wang, ``Semantic correlation
  promoted shape-variant context for segmentation,'' in \emph{Proc. IEEE Conf.
  Comput. Vis. Pattern Recognit.}, 2019, pp. 8885--8894.

\bibitem{BFP}
H.~Ding, X.~Jiang, A.~Q. Liu, N.~M. Thalmann, and G.~Wang, ``Boundary-aware
  feature propagation for scene segmentation,'' in \emph{Proc. IEEE Int. Conf.
  Comput. Vis.}, 2019, pp. 6819--6829.

\bibitem{ding2023self}
H.~Ding, H.~Zhang, and X.~Jiang, ``Self-regularized prototypical network for
  few-shot semantic segmentation,'' \emph{Pattern Recognition}, vol. 133, p.
  109018, 2023.

\bibitem{liu2021few}
W.~Liu, Z.~Wu, H.~Ding, F.~Liu, J.~Lin, and G.~Lin, ``Few-shot segmentation
  with global and local contrastive learning,'' \emph{arXiv preprint
  arXiv:2108.05293}, 2021.

\bibitem{OSLSM}
A.~Shaban, S.~Bansal, Z.~Liu, I.~Essa, and B.~Boots, ``One-shot learning for
  semantic segmentation,'' in \emph{British Machine Vision Conference}.\hskip
  1em plus 0.5em minus 0.4em\relax {BMVA} Press, 2017.

\bibitem{koch2015siamese}
G.~Koch, R.~Zemel, and R.~Salakhutdinov, ``Siamese neural networks for one-shot
  image recognition,'' in \emph{ICML deep learning workshop}, vol.~2.\hskip 1em
  plus 0.5em minus 0.4em\relax Lille, 2015.

\bibitem{dong2018few}
N.~Dong and E.~Xing, ``Few-shot semantic segmentation with prototype
  learning.'' in \emph{Proceedings of the British Machine Vision Conference,
  Newcastle, UK}.\hskip 1em plus 0.5em minus 0.4em\relax {BMVA} Press, 2018,
  p.~79.

\bibitem{SG-one}
X.~Zhang, Y.~Wei, Y.~Yang, and T.~S. Huang, ``Sg-one: Similarity guidance
  network for one-shot semantic segmentation,'' \emph{{IEEE} Trans. Cybern.},
  vol.~50, no.~9, pp. 3855--3865, 2020.

\bibitem{zhang2019canet}
C.~Zhang, G.~Lin, F.~Liu, R.~Yao, and C.~Shen, ``Canet: Class-agnostic
  segmentation networks with iterative refinement and attentive few-shot
  learning,'' in \emph{Proc. IEEE Conf. Comput. Vis. Pattern Recognit.}, 2019,
  pp. 5217--5226.

\bibitem{tian2020pfenet}
Z.~Tian, H.~Zhao, M.~Shu, Z.~Yang, R.~Li, and J.~Jia, ``Prior guided feature
  enrichment network for few-shot segmentation,'' \emph{{IEEE} Trans. Pattern
  Anal. Mach. Intell.}, vol.~44, no.~2, pp. 1050--1065, 2022.

\bibitem{chu2016structured}
X.~Chu, W.~Ouyang, H.~Li, and X.~Wang, ``Structured feature learning for pose
  estimation,'' in \emph{Proc. IEEE Conf. Comput. Vis. Pattern Recognit.},
  2016, pp. 4715--4723.

\bibitem{word2vector}
T.~Mikolov, K.~Chen, G.~Corrado, and J.~Dean, ``Efficient estimation of word
  representations in vector space,'' in \emph{ICLR}, 2013.

\bibitem{clip}
A.~Radford, J.~W. Kim, C.~Hallacy, A.~Ramesh, G.~Goh, S.~Agarwal, G.~Sastry,
  A.~Askell, P.~Mishkin, J.~Clark \emph{et~al.}, ``Learning transferable visual
  models from natural language supervision,'' in \emph{Proc. Int. Conf. Mach.
  Learn.}\hskip 1em plus 0.5em minus 0.4em\relax PMLR, 2021, pp. 8748--8763.

\bibitem{ding2021vision}
H.~Ding, C.~Liu, S.~Wang, and X.~Jiang, ``Vision-language transformer and query
  generation for referring segmentation,'' in \emph{ICCV}, 2021, pp.
  16\,321--16\,330.

\bibitem{ding2020phraseclick}
H.~Ding, S.~Cohen, B.~Price, and X.~Jiang, ``Phraseclick: toward achieving
  flexible interactive segmentation by phrase and click,'' in
  \emph{ECCV}.\hskip 1em plus 0.5em minus 0.4em\relax Springer, 2020, pp.
  417--435.

\bibitem{VLTPAMI}
H.~Ding, C.~Liu, S.~Wang, and X.~Jiang, ``{VLT:} vision-language transformer
  and query generation for referring segmentation,'' \emph{IEEE TPAMI},
  vol.~45, no.~6, pp. 7900--7916, 2023.

\bibitem{GMMN}
Y.~Li, K.~Swersky, and R.~Zemel, ``Generative moment matching networks,'' in
  \emph{Proc. Int. Conf. Mach. Learn.}\hskip 1em plus 0.5em minus 0.4em\relax
  PMLR, 2015, pp. 1718--1727.

\bibitem{armeni20163d}
I.~Armeni, O.~Sener, A.~R. Zamir, H.~Jiang, I.~Brilakis, M.~Fischer, and
  S.~Savarese, ``3d semantic parsing of large-scale indoor spaces,'' in
  \emph{Proc. IEEE Conf. Comput. Vis. Pattern Recognit.}, 2016, pp. 1534--1543.

\bibitem{dai2017scannet}
A.~Dai, A.~X. Chang, M.~Savva, M.~Halber, T.~Funkhouser, and M.~Niessner,
  ``Scannet: Richly-annotated 3d reconstructions of indoor scenes,'' in
  \emph{Proc. IEEE Conf. Comput. Vis. Pattern Recognit.}, 2017, pp. 5828--5839.

\bibitem{CWT}
Z.~Lu, S.~He, X.~Zhu, L.~Zhang, Y.-Z. Song, and T.~Xiang, ``Simpler is better:
  Few-shot semantic segmentation with classifier weight transformer,'' in
  \emph{ICCV}, 2021, pp. 8741--8750.

\bibitem{detr}
N.~Carion, F.~Massa, G.~Synnaeve, N.~Usunier, A.~Kirillov, and S.~Zagoruyko,
  ``End-to-end object detection with transformers,'' in \emph{ECCV}.\hskip 1em
  plus 0.5em minus 0.4em\relax Cham: Springer International Publishing, 2020,
  pp. 213--229.

\bibitem{garcia2017few}
V.~Garcia and J.~Bruna, ``Few-shot learning with graph neural networks,'' in
  \emph{6th International Conference on Learning Representations, Vancouver,
  BC, Canada, Conference Track Proceedings}.\hskip 1em plus 0.5em minus
  0.4em\relax OpenReview.net, 2018.

\bibitem{FPN}
T.-Y. Lin, P.~Doll{\'a}r, R.~Girshick, K.~He, B.~Hariharan, and S.~Belongie,
  ``Feature pyramid networks for object detection,'' in \emph{Proc. IEEE Conf.
  Comput. Vis. Pattern Recognit.}, 2017, pp. 2117--2125.

\bibitem{CGBNet}
H.~Ding, X.~Jiang, B.~Shuai, A.~Q. Liu, and G.~Wang, ``Semantic segmentation
  with context encoding and multi-path decoding,'' \emph{IEEE Transactions on
  Image Processing}, vol.~29, pp. 3520--3533, 2020.

\bibitem{zhang2021pointclip}
R.~Zhang, Z.~Guo, W.~Zhang, K.~Li, X.~Miao, B.~Cui, Y.~Qiao, P.~Gao, and H.~Li,
  ``Pointclip: Point cloud understanding by clip,'' in \emph{Proc. IEEE Conf.
  Comput. Vis. Pattern Recognit.}, 2022, pp. 8552--8562.

\bibitem{3DGenZ}
B.~Michele, A.~Boulch, G.~Puy, M.~Bucher, and R.~Marlet, ``Generative zero-shot
  learning for semantic segmentation of 3d point clouds,'' in \emph{2021
  International Conference on 3D Vision (3DV)}.\hskip 1em plus 0.5em minus
  0.4em\relax IEEE, 2021, pp. 992--1002.

\end{thebibliography}
}

\end{document}